\documentclass{article}

\usepackage{arxiv}

\usepackage[utf8]{inputenc} 
\usepackage[T1]{fontenc}    
\usepackage{hyperref}       
\usepackage{url}            
\usepackage{booktabs}       
\usepackage{amsfonts}       
\usepackage{nicefrac}       
\usepackage{microtype}      
\usepackage{graphicx}
\usepackage{doi}
\usepackage{subcaption}
\usepackage{amsmath, amssymb, amsfonts}

\DeclareMathOperator*{\argmin}{arg\,min}
\DeclareMathOperator*{\argmax}{arg\,max}

\title{Deep Canonical Correlation Alignment for Sensor Signals}

\author{ 
    \href{https://orcid.org/0000-0001-5069-407X}{\includegraphics[scale=0.06]{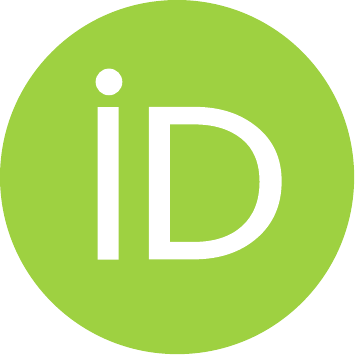}\hspace{1mm}Narayan Schütz}\thanks{contributed equally to this work} \\
	ARTORG Center for Biomedical Research\\
	University of Bern\\
	3008 Bern,  Switzerland\\
	\texttt{narayan.schuetz@artorg.unibe.ch} \\
	\And
	\href{https://orcid.org/0000-0002-1245-2931}{\includegraphics[scale=0.06]{orcid.pdf}\hspace{1mm}Angela Botros}$^*$ \\
	ARTORG Center for Biomedical Research\\
	University of Bern\\
	3008 Bern,  Switzerland\\
	\texttt{angela.botros@artorg.unibe.ch} \\
	\And
	\href{https://orcid.org/0000-0002-4883-9484}{\includegraphics[scale=0.06]{orcid.pdf}\hspace{1mm}Michael Single}$^*$ \\
	ARTORG Center for Biomedical Research\\
	University of Bern\\
	3008 Bern,  Switzerland\\
	\texttt{michael.single@artorg.unibe.ch} \\
	\And
	\href{https://orcid.org/(0000-0002-1797-9441}{\includegraphics[scale=0.06]{orcid.pdf}\hspace{1mm}Aileen C. Naef} \\
	ARTORG Center for Biomedical Research\\
	University of Bern\\
	3008 Bern,  Switzerland\\
	\texttt{aileen.naef@artorg.unibe.ch} \\
	\And
	\href{https://orcid.org/(0000-0001-6577-200X}{\includegraphics[scale=0.06]{orcid.pdf}\hspace{1mm}Philipp Buluschek} \\
	DomoSafety S.A.\\
	EPFL Innovation Park, Building D\\
	1015 Lausanne,  Switzerland\\
	\texttt{philipp.buluschek@domo-safety.com} \\
	\And
	\href{https://orcid.org/0000-0002-8069-9450}{\includegraphics[scale=0.06]{orcid.pdf}\hspace{1mm}Tobias Nef} \\
	ARTORG Center for Biomedical Research\\
	University of Bern\\
	3008 Bern,  Switzerland\\
	\texttt{tobias.nef@artorg.unibe.ch}
}

\renewcommand{\shorttitle}{DCCA for Sensor Signals}

\hypersetup{
pdftitle={Deep Canonical Correlation Alignment for Sensor Signals},
pdfsubject={q-bio.NC, q-bio.QM},
pdfauthor={Narayan Schütz, Angela Botros, Michael Single, Tobias Nef},
pdfkeywords={Biomedical signal processing, Machine learning algorithms, Cross Correlation. Canonical Correlation, Signal Alignment, Sensors},
}

\begin{document}
\maketitle

\begin{abstract}
Sensor technologies are becoming increasingly prevalent in the biomedical field, with applications ranging from telemonitoring of people at risk, to using sensor derived information as objective endpoints in clinical trials.
To fully utilize sensor information, signals from distinct sensors often have to be temporally aligned.
However, due to imperfect oscillators and significant noise, commonly encountered with biomedical signals, temporal alignment of raw signals is an all but trivial problem, with, to-date, no generally applicable solution.
In this work, we present Deep Canonical Correlation Alignment (DCCA), a novel, generally applicable solution for the temporal alignment of raw (biomedical) sensor signals.
DCCA allows practitioners to directly align raw signals, from distinct sensors, without requiring deep domain knowledge.
On a selection of artificial and real datasets, we demonstrate the performance and utility of DCCA under a variety of conditions.
We compare the DCCA algorithm to other warping based methods, DCCA outperforms dynamic time warping and cross correlation based methods by an order of magnitude in terms of alignment error.
DCCA performs especially well on almost periodic biomedical signals such as heart-beats and breathing patterns.
In comparison to existing approaches, that are not tailored towards raw sensor data, DCCA is not only fast enough to work on signals with billions of data points but also provides automatic filtering and transformation functionalities, allowing it to deal with very noisy and even morphologically distinct signals.
\end{abstract}


\section{Introduction}
\label{sec:introduction}
With the rise of the internet of things (IoT), the amount of sensor generated data is growing exponentially \cite{ahmed2017role}. 
The biomedical field is particularly affected by this trend with an increasing digitalization in medicine and a flood of new biomedical sensing devices being used to monitor patients in, and more frequently, also outside the clinics. 
As such, sensor derived information is being used to evaluate patient health status, trigger alarms in emergency situations, and to develop objective digital endpoints in clinical trials \cite{rantz2015new, saner2021case, schutz2021contactless, evidation2019, amin2016radar, boehme2019soon, coravos2020modernizing}.
Moreover, biomedical sensors are becoming ubiquitous in consumer devices, like smart watches or fitness devices that are often used in recreational, and more frequently, also in a variety of research settings \cite{henriksen2018using, king2018survey, shin2019wearable}.\\
\newline
With the increase in sensor data, one particular issue that is becoming more apparent, is accurate temporal alignment \cite{bent2020investigating, jiang2020eventdtw, vollmer2019alignment, schutz2020real}.
There can be many reasons contributing to temporal misalignments between multiple sensors, with the most obvious being imperfect sampling frequencies.
At its core, this problem cannot currently be mitigated, as most practical oscillators such as quartz crystals, used for frequency control, will exhibit some inaccuracy over time.
This may be amplified by external factors like temperature or varying battery levels \cite{zhou2008frequency}.
As a result, the exhibited sampling frequencies will always be subject to small, potentially non-linear, inaccuracies, often referred to as offset and clock-drift \cite{sivrikaya2004time}.
This phenomenon is especially noticeable with sensors having high ($\gg$ 1Hz) sampling frequencies as often encountered in biomedical settings.
In pre-designed multi-sensor networks, this may be corrected for, by using specialized synchronization algorithms, working in real-time or close to real-time \cite{sivrikaya2004time, lasassmeh2010time}. 
However, in many realistic scenarios, the overhead of such synchronisation techniques would be cost ineffective, impractical, or not accounted for, at the time of acquisition. 
This poses significant problems, especially in the biomedical field.
One such problem is the difficulty regarding comparison, verification and validation of signals and algorithms against clinically validated gold-standards.
Another problem is the fusion of information from different sensing devices. 
Of particular interest, especially in the context of increasing artificial intelligence adoption, is the transfer of algorithms and knowledge between devices.
An example would be the transfer of R-peak extraction algorithms from a validated electrocardography (ECG) device to a new photoplethysmography (PPG) based device;
or transferring activity recognition algorithms from one sensor and body-position to another one, at a slightly different position.
Oftentimes, a significant amount of effort has to be put into creating new, manually annotated, and curated, datasets for each new variation, even if this may just be a different sensor of the same type.
Given accurate temporal alignment of raw signals, these situations could easily be turned into self-supervised learning scenarios, where minimal human effort is necessary to transform knowledge and algorithms from one device to another, as well as validate the same.
A particular advantage in biomedical settings is the fact that we are commonly measuring humans.
As such, sensor signals are very often inter-correlated to a certain extent, even if they were obtained by measuring different physical parameters.
Take for instance an electroencephalogram (EEG) and an ECG signal:
One measures the electrical activity of the brain, the other one the electrical activity of the heart. 
In theory, these two should be mostly uncorrelated processes.
However, due to the electrical activity of the heart, cardiac field artifacts can be measured in EEG signals \cite{harke1999cardiac}.
As a result, the signals may still be aligned on the basis of such indirect effects, given the signals are suitably pre-processed.
Such indirect correlations are very common within human recordings and can be exploited to align even very different sensor modalities.\\

At the time of writing, there is no generally applicable solution to temporally align long, inter-correlated, raw sensor signals that may be morphologically distinct. 
Most approaches usually rely on significant pre-processing of the raw sensor data (such as filtering, transforming and denoising).
Oftentimes, this includes the extraction of certain high-level features, on top of which the alignment is then performed \cite{vollmer2019alignment, huysmans2019evaluation}. 
Traditional approaches to temporally align time-series data, such as sensor signals, often involve time-delay analysis using Cross-Correlation (CC) or Dynamic Time Warping (DTW) based approaches \cite{rhudy2014time, jiang2020eventdtw}. 
These approaches work well for shorter, usually already pre-processed signals but are less suited towards longer (millions to billions of data points) raw sensor signals. 
While CC based approaches are naturally limited by their inability to incorporate warping (such as would be introduced by clock-drift and offsets), DTW approaches tend to collapse to single points, disregarding the continuous shift dynamic of sensors \cite{zhang2017dynamic}. 
In theory, some of those downsides can be individually addressed, for instance, by penalizing the number of links in DTW \cite{zhang2017dynamic}, using approximate solutions \cite{salvador2007toward}, or by using piece-wise warping functions, for example by Windowed Cross-Correlation (WCC) \cite{schutz2020real}.
However, even if the above mentioned issues are addressed, we are still not referring to raw, potentially very long and morphologically distinct sensor signals but assume already cleaned and pre-processed signals. 
To get to this stage, usually, deep domain knowledge and expertise in advanced digital signal processing is required, as each new signal may come with its own special cases and unique requirements.\\

In this work, we share a general temporal signal-alignment algorithm that we have been using to automatically align thousands of hours of real-world biomedical sensor signals, amounting to billions of data points from more than one hundred distinct sensors. 
The presented approach is specifically tailored to perform automatic temporal alignment of raw, noisy, potentially multivariate, and morphologically distinct inter-correlated sensor signals that were sampled with high sampling frequencies. 

Our main contributions are:
\begin{itemize}
    \item Introduction of a, novel, fast and general temporal alignment algorithm that automatically aligns raw sensor signals, which may be multivariate, very long (millions - billions of data points), highly noisy and morphologically distinct. 
    Something, no current method manages to achieve. 
    \item Thorough evaluation of the proposed algorithm on a variety of publicly available synthetic and real-world datasets.
    \item Demonstration of the approach on real-world alignment scenarios.
\end{itemize}

While we focus on biomedical sensor signals, it is reasonable to assume that the proposed approach may also be similarly used in other fields, especially as sensor technology is becoming more and more ubiquitous.

\section{Problem Definition}\label{section:problem_definition}
\begin{figure}
\centering
\includegraphics[width=0.7\columnwidth]{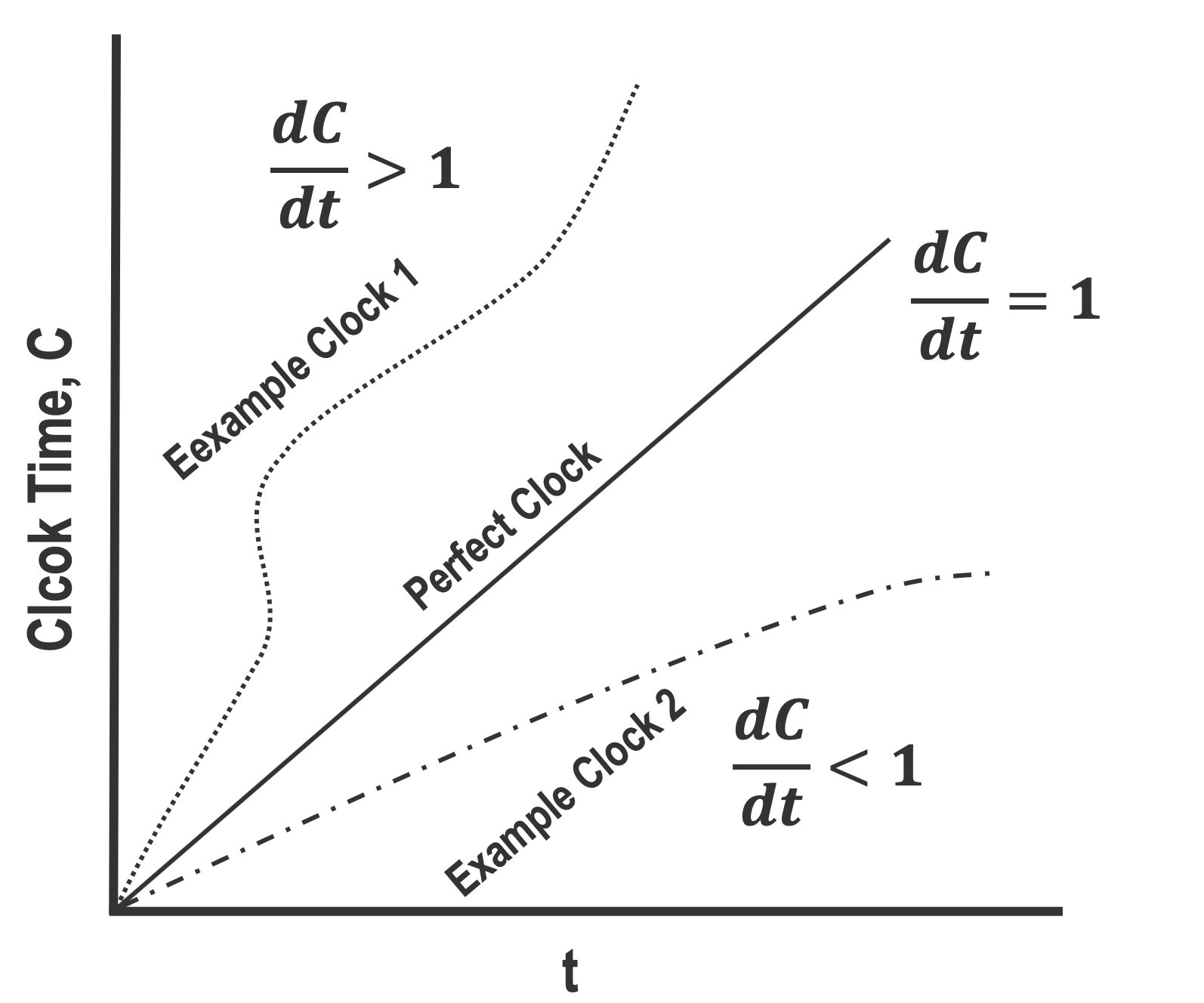}%
\caption{Intuitive example of clock-drift on the basis of real clocks.
Due to imperfect oscillators, real clocks tend to be faster or slower, compared to a perfect clock. 
Depending on the oscillator quality this behavior may be non-linear. 
The same applies to the oscillators used for frequency control in most modern sensors}
\label{fig:clockdrift}
\end{figure}

An intuitive example of our problem is the measurement of time using two different regular clocks.
Here, the same event, time, is measured using two different independent sensors, the clocks.
As can be seen in the schematic of Fig.~\ref{fig:clockdrift}, the different derivatives of clock-time with respect to time $\frac{\partial C}{\partial t}$ lead to growing displacements between the two clocks.
While for higher-quality oscillators $\frac{\partial C}{\partial t}$ can be assumed to be a constant value close to 1, it may be subject to some small changes due to slow oscillations of $\frac{\partial^2 C}{\partial t^2}$ around 0.
These are commonly the result of external influences like changes in temperature or battery levels~\cite{vollmer2019alignment}.
Additionally to clock-drift, we often encountered small instantaneous jumps, further referred to as offsets, between many real-world sensor pairs - especially with less expensive devices.
While we are not exactly certain of their origin, we assume those to be related to short-term sensor failures, internal synchronization attempts or connection failures, leading to the apparent loss of a smaller number of samples.
One particularity to note, that we will make use of later on, is that oscillators used in modern sensors, even in cheaper ones, are generally not wildly inaccurate.
Thus, it is safe to assume that $\frac{\partial C}{\partial t} \approx 1$ and $\frac{\partial^2 C}{\partial t^2} \approx 0$ over shorter time intervals of few seconds to few minutes.\\

To correct for the above described displacements, we are looking for a continuous warping function that characterises displacements at each time $t$ of a signal recording. 
Formally, our goal is thus to extract a warping function
\begin{align}
d(t) &= h(t) + b(t),
\label{eq:problem}
\end{align}
where $h(\cdot)$ is a continuous function, modelling the clock-drift between two sensors at time $t$, and $b(\cdot)$ a step function, representing instantaneous offsets at time $t$.
Approximating $d(\cdot)$ by a function $\hat{d}(\cdot)$ would thus allow us to correct temporal misalignments between a given pair of signals. 

Since we intend to align raw, very long, biomedical time-series signals, there are some requirements that should be met by an algorithm that approximates $d(t)$: 1) the algorithm needs to be fast enough to be viable in the real-world; 2) the algorithm needs to be able to deal with severe noise; 3) the algorithm has to work with partially correlated signals; 4) the algorithm needs to be based on local similarities as opposed to global ones; 5) the algorithm should be able to deal with morphologically distinct signals; 6) $\hat{d}(t)$ should be able to accommodate nonlinear behavior to a certain degree. 

While the first requirement does not need further elucidation given many sensors sample at frequencies far above 1 Hz and record for hours on end, we will go into some further detail with regards to the other requirements and give examples in the biomedical context. \newline
Beginning with requirement 2), many real world biomedical signals often contain severe noise coming in a variety of types, e.g. movement artifacts, power line interferences, signal losses, and inherent measurement inaccuracies.
Requirement 3) sometimes coincides with 2), in that noise can lead to partially uncorrelated signals. 
However, in some scenarios it may be inherent to the respective sensors or measurement set-up that the corresponding signals are only partially correlated.
One example would be two accelerometers worn on the wrist and on the ankle, where only a subset of all activities can be recorded by both sensors - such as walking.
Coming to requirement 4), it is important to realise that raw biomedical signals are oftentimes almost periodic~\cite{trajkovic2009modelling} but are subject to a variety of external perturbations, most notably movements of all kinds.
This may lead to phenomena such as baseline wander, commonly encountered within ECG recordings.
As a consequence, global structure may be dominated by uncorrelated noise and cannot be used for alignment if one intends to work with unfiltered signals. 
Local structure, on the other hand, such as smaller variations in periodic behavior (think of heart rate or respiration rate variability), are often what allows one to detect the correct shift between two signals. 
Requirement 5) is highly important when dealing with different sensor types that measure the same event. 
Examples of this, related to biomedical signals, are the aforementioned measurement of heartbeats using ECG, ballistocardiography (BCG), PPG - and many more potential modalities like radar, sound or skin temperature.
Finally, requirement 6) follows from the observation that clock-drift can be non-linear~\cite{vollmer2019alignment}, thus at least for larger $\Delta t$, a suitable warping function should be able to approximate non-linear behavior.

\section{Related Work}\label{sec:related_work}
Assume we are given two real valued similar functions $S_1(t), S_2(t)$ such that
\begin{equation}
    S_1(t) \approx S_2( d(t) ) + \sigma(t). \label{eq:problemdescription}
\end{equation}
These two signals are similar but not identical in value, and the time of $S_2$ is somehow warped by an unknown function $d(t)$.
Additionally, there is some random noise $\sigma(t)$.
Depending on the shape of the time warping function $d(t)$, there are two primary categories of approaches, when trying to align two signals.

\subsection{Constant Warping}
The first category covers approaches based on Cross-Correlation (CC).
These have been used in a multitude of practical applications \cite{rhudy2014time, zhou2009canonical, zhou2012generalized}.
CC related approaches make the assumption that the underlying warping function is linear and characterized by a constant time shift $c$, thus of the form $d_{CC}(t) = t + c$.
Formally, given two time signals $S_1, S_2 \in \mathbb{R}^N$, the problem can be described as
\begin{equation}
    S_1(t) \approx S_2(t+c) + \sigma(t). \label{eq:CC_basis}
\end{equation}
An approximation $\hat{c}$ can be calculated by solving
\begin{equation}
     \hat{c} = \argmin\limits_{c} \sum_k \left(S_1 [ k-c ] - S_2 [ k ] \right)^2.   \label{eq:minshift}
\end{equation}
This approximation $\hat{c}$ can, for instance, be found with Normalized Cross-Correlation (NCC).
The NCC between two signals is defined as
\begin{equation}\begin{array}{rl}
    NCC(S_1, S_2) &=\Big \langle \frac{S_1}{\|S_1\|_F}, \frac{S_2}{\|S_2\|_F}\Big \rangle \vspace{0.2cm}\\
        &= (S_1)^{-1/2}S_1S_2^T(S_2^T)^{-1/2}.
        \end{array}
        \label{eq:NCC_vec}
\end{equation}
To find an approximation of the time shift $c$, the NCC can be optimized by solving
\begin{equation}
    \begin{array}{rl}
         \argmax\limits_{c} & \Big \langle \dfrac{S_1\Delta_1}{\|S_1\Delta_1\|}, \dfrac{S_2\Delta_2}{\|S_2\Delta_2\|}\Big \rangle \\
         \textrm{subject to} &  \Delta_1 \in \mathcal{D}_1(c) \\
                       & \Delta_2 \in \mathcal{D}_2(c)
    \end{array} \label{eq:NCC_opt}
\end{equation}

where the sets $\mathcal{D}_1(c)$ and $\mathcal{D}_2(c)$ are defined as
\begin{equation}\label{eq:objective_cc}
    \begin{array}{l}
    \mathcal{D}_1(c): \Delta[m,n] = \left\{ \begin{array}{rl}
        1 & \Delta[m,m+c] = 1 \\
        0 & else
    \end{array}\right. \\
    \mathcal{D}_2(c): \Delta[m,n] = \left\{ \begin{array}{rl}
        1 & \begin{array}{l}
            \max(0,-c) \leq m=n  \\
            m\leq \min(N-c,N)  
        \end{array}\\
        0 & else
    \end{array}\right.
    \end{array}
\end{equation}
The matrices $\Delta_1, \Delta_2 \in \{0,1\}^{N\times N}$ are sparse and ensure the correct slicing of the signals $S_1$ and $S_2$ as given in~\eqref{eq:minshift}. 
$\Delta_1$ is the $c$-off-diagonal matrix and $\Delta_2$ is a diagonal matrix with only nonzero entries on the diagonal elements that are overlapping signal elements.
There is only one degree of freedom in~\eqref{eq:NCC_opt}, the parameter $c$. \newline
Equation~\eqref{eq:NCC_opt} can be formulated as a least-squares problem
\begin{equation}
    \begin{array}{rl}
         \argmin\limits_{c} & \|S_1\Delta_1 - S_2\Delta_2 \|^2_F \\
         \textrm{subject to} &  \Delta_1 \in \mathcal{D}_1(c) \vspace{0.1cm}\\
                       & \Delta_2 \in \mathcal{D}_2(c).
    \end{array} \label{eq:NCC_LeastSquares}
\end{equation}

\subsection{Non-Linear Warping}
The second category of approaches is based on the Dynamic Time Warping (DTW) distance measure~\cite{rabiner1993fundamentals}, which assumes the warping function $d_{DTW}(t)$ between two signals to be continuous and non-linear with certain constraints.
With DTW based approaches, two signals $S_1$ and $S_2$ are aligned using the DTW distance measure by solving

\begin{equation}\label{eq:DTW}
\begin{array}{rl}
    \argmin\limits_{\Delta_1, \Delta_2}  & \|S_1\Delta_1 - S_2\Delta_2\|_F^2 \\
     \textrm{subject to} & \Delta_i \in \{0,1\}^{N\times N},~~ i\in{1,2}. 
\end{array} 
\end{equation}

In~\cite{mueen_extracting_2016}, a comprehensive tutorial on DTW is provided. 
It is possible that elements of one signal match with multiple elements of the other signal and vice versa.
This method works as long as the warping function is monotonously increasing, i.e. there are no jumps back in time.
Compared to NCC, which optimizes exactly one parameter, the constant time shift $c$, DTW has many more degrees of freedom.
It should be noted that there are many DTW flavors, and with EventDTW even a variant specifically designed to align (pre-processed) sensor signals exhibiting different sampling frequencies \cite{jiang2020eventdtw}.
However, they all work by the same principle described in~\eqref{eq:DTW}, with ad-hoc modifications in constraints (mainly windows and step patterns) or specific pre-processing of the input signal.\\
Additionally to exact DTWs, there are more computationally efficient inexact approaches, like FastDTW \cite{delong2012fast} or generalized time warping (GTW) \cite{zhou2012generalized}, that can reduce time and space complexity from to $\mathcal{O}(n^2)$ to $\mathcal{O}(n)$, at the cost of a non-exact alignments.

\subsection{Linear Warping}
Finally, it is important to note that there are cases in-between those two major categories of warping functions. 
For instance, linear warping, where $d(t)$ models an affine transformation and is thus of the form $d(t)_{affine} = bt + c$ \cite{williams2020discovering}. 
Additionally, there are approximations to the non-linear form, such as piece-wise linear warping functions, which can be estimated with WCC as we made use of in \cite{schutz2020real}. 

\subsection{Extension to Multivariate Alignments}
To calculate linear relationships between multivariate time signals (with potentially different dimensionalities), Canonical Correlation Analysis (CCA)~\cite{hotelling1992relations} may be used.
With CCA two multivariate time signals $S_i \in \mathbb{R}^{l_i \times N}$ with  $i = \{1,2\}$ are projected to a common (lower) dimensional sub space using $W_i \in \mathbb{R}^{l_i\times l}$ such that $Y_i = W_i^TS_i$ with  $Y_i \in \mathbb{R}^{l \times N}$.
The best projection is obtained by minimizing
\begin{equation}
\begin{array}{rl}
    \argmin\limits_{W_1, W_2} & \| W_1^TS_1 - W_2^TS_2 \|^2_F \\ 
    \textrm{subject to} & W_i^TS_iS_i^TW_i = I, ~~ i=\{1,2\}
\end{array} \label{eq:CCA}
\end{equation}
In order to find the optimal alignment of two multivariate time signals, NCC from~\eqref{eq:NCC_LeastSquares} can be combined with CCA.
\begin{equation}
    \begin{array}{rl}
         \argmin\limits_{W_1, W_2,d} & \|W_1^TS_1\Delta_1 - W_2^TS_2\Delta_2 \|_F^2 \\
         \textrm{subject to} &  W_i^TS_i\Delta_i\Delta_i^TS_i^TW_i = I, ~~ i \in \{1,2\} \vspace{0.1cm}\\
                       & \Delta_1 \in \mathcal{D}_1(c) \\
                       & \Delta_2 \in \mathcal{D}_2(c)
    \end{array} \label{eq:NCC_CCA}
\end{equation}

Similarly, DTW can be extended with CCA to accommodate different input dimensionalities, referred to as Canonical Time Warping (CTW) \cite{zhou_canonical_nodate}. 
\begin{equation}
    \begin{array}{rl}
         \argmin\limits_{W_1, W_2, \Delta_1, \Delta_2} & \|W_1^TS_1\Delta_1 - W_2^TS_2\Delta_2 \|_F^2 \\
         \textrm{subject to} &  W_i^TS_i\Delta_i\Delta^T_iS_i^TW_i = I,~~ i \in \{1,2\} \vspace{0.1cm}\\
                       & W_1^TS_1\Delta_1\Delta^T_2S_2^TW_2 = D \vspace{0.1cm}\\
                       & \Delta_i \in \{0,1\}^{l_i\times l}, ~~ i \in \{1,2\}
    \end{array} \label{eq:CTW}
\end{equation}
In~\cite{trigeorgis_deep_2015}, CTW is expanded to Deep Canonical Time Warping (DCTW) by replacing the linear mapping $W_i^TS_i$ by a non-linear function $f(S_1; \theta_1)$ that is approximated by a neural network.

\section{Deep Canonical Correlation Alignment}
To the best of our knowledge, no algorithm that fulfills all requirements stated in \ref{section:problem_definition} exists.
Constant and linear warping functions are clearly not enough to fulfill the requirements of potentially non-linear clock-drift and offsets. 
Non-linear warping approaches such as exact DTWs are not computationally efficient enough for very long sensor signals.
Furthermore, they have no obvious mechanism to deal with partially correlated signals, let alone strong morphological differences along the time-domain.
Additionally, we found in preliminary experiments that non-linear warping function based approaches are too general (thus have too many degrees of freedom, in the context of selection matrices $\Delta$.) to reliably deal with almost periodic biomedical signals.
In other words, non-linear warping tends to "overfit" by eliminating the small inter-cycle variabilities, which are exactly what would allow for a correct alignment to be found.
For the interested reader, we provide an example and further information with regards to this behavior in the supplementary material.

Based on these observations it makes sense to use a slightly less powerful warping function, thus a piece-wise approximation of non-linear warping functions.
One way to achieve this is by means of WCC that can be used to successfully align properly pre-processed, raw, biomedical signals \cite{schutz2020real}.
However, WCC and comparable piece-wise linear approaches are not able to deal with severe noise, partially correlated signals, and morphologically distinct signals, our requirements 2) - 5). \\

To address this, we introduce Deep Canonical Correlation Alignment (DCCA).
DCCA extends WCC in multiple ways.
First it extends NCC with CCA, to allow for working with multi-variate signals with, potentially, different dimensions.
Second, the projection functions in CCA are extended with non-linear transformation functions, similar to DCTW, that can be approximated by deep neural networks.
However, since we work on discrete windows of points, instead of individual points as in DCTW, our non-linear function is much more powerful, allowing mappings of the form $\mathbb{R}^{l_i \times T_w} \rightarrow \mathbb{R}^{l_j \times T_w}$, instead of $\mathbb{R}^{l_i} \rightarrow \mathbb{R}^{l_j}$, thus considering time-dynamics of window $T_w$ and not only the feature space $\mathbb{R}^{l_i}$.
This step allows to both filter and morph signal pairs to a maximally correlated latent space, where the local shift can be more reliably estimated.
Lastly, DCCA extracts knot points, defining the piece-wise approximation, and uses them for robust, energy-based multi-model extraction.
This is in contrast to the naive approach of knot points directly defining a piece-wise linear approximation. 
As a result, DCCA is able to extract more sophisticated piece-wise warping functions (such as piece-wise polynomial), while correcting for the influence of noise and partially correlated signals.
Additionally, this also allows for the incorporation of hard constraints on the desired warping function, such as for instance a limit on curvature.
Finally, a nice side-effect of using an alignment approach based on NCC and CCA, is the possibility to completely vectorize this step and express it with highly optimized tensor operations, which may be calculated on specialized hardware like GPUs or tensor processing units.

\subsection{Segmentation}
\begin{figure}
\centering
\includegraphics[width=0.5\columnwidth]{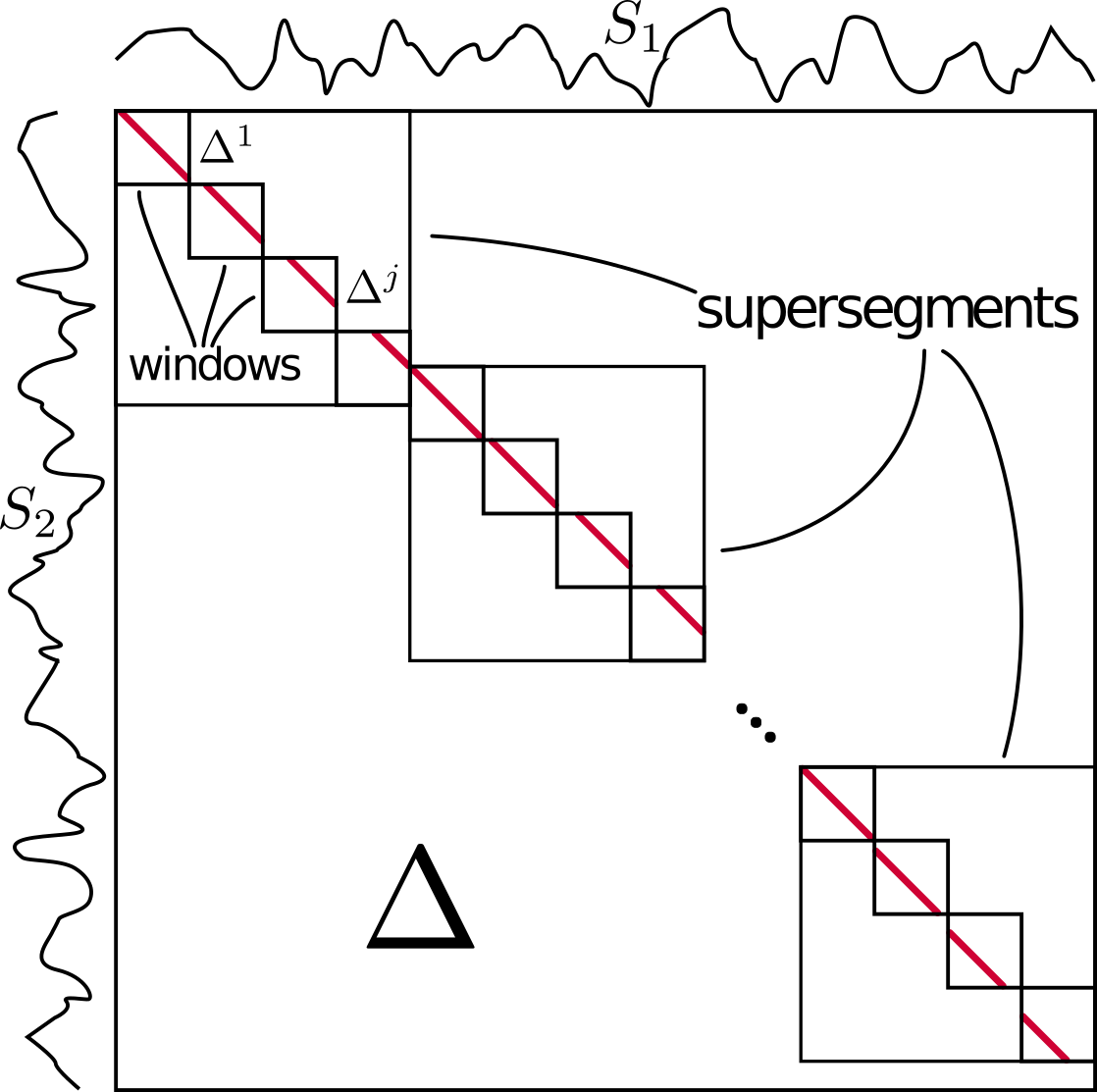}
\caption{This is a representation of the complete alignment matrix $\Delta$, matching $S_1$ to $S_2$. $\Delta$ is subdivided into the super-segments. Every super-segment is subdivided into windows. The red lines denotes the true alignment.}
\label{fig:alignmentsquares}
\end{figure}

As we aim to find a piece-wise approximation of warping function $d(\cdot)$, we need to evaluate $d(\cdot)$ at a number of knot points $\tau_1,...,\tau_M$.
One way to do this is to segment the signals into small discrete windows $w$ that can be used to find estimates of $d(\cdot)$ around a given knot point $\tau_j$.
Using a small window to evaluate $d(\cdot)$ at a given $\tau_j$, instead of actual point estimates, works because of our assumption on warping functions generated by modern sensors.
Namely, that they change very slowly, except for the occasional offset.
Given the two signals $S_{1} \in \mathbb{R}^{l_{1}\times N}$ and $S_{2} \in \mathbb{R}^{l_{2}\times N}$, we first split them into $Z = \frac{N}{z}$ super-segments of length $z$.
Each of the $Z$ super-segments is further split into $M = \frac{Z}{w}$ windows of length $w$. 
A schematic of this segmentation is illustrated in Fig.~\ref{fig:alignmentsquares} and in the first part of Fig.~\ref{fig:schematic}.
For window length $w$, we assume the time shift difference of the $k$th window is
$\Delta d_k = d(w\cdot(k+1)) - d(w\cdot k) \approx 0$.
This also gives an upper limit on the chosen window length $w$ as it limits the steepness of the time warping function.
A lower bound on the window length is posed by the largest time shift over the whole signal $\max(|d(t) - t|) \leq \lambda w, ~~ \lambda \in (0, 1)$.
The window length $w$ should thus be larger than the maximum time shift, as otherwise some windows will have no signal overlap.
This is where super segments become helpful, as they allow one to keep $w$ shorter than $\max(d(t))$.
As a consequence, super segment length $z$ must be chosen such that the maximum displacement within itself is smaller than what can maximally be contained within $w$.
Further considerations on how to set $z$ and $w$, as well as reasonable default values, are reported in the supplementary material.
For all further practical applications, we manually set $\lambda=0.5$.
These assumptions and settings allow us to treat every signal segment pair as if $d(t)$ would be linearized and~\eqref{eq:CC_basis} would hold.

While windows $M$ are completely independent, super-segments $Z$ can be mostly treated independently as well, if the sensors are connected to the internet and perform some kind of network protocol requests - which provides start points for alignment.
In these cases calculations within segments and super-segments are completely parallelizable.
However, if it is not the case, super-segments can be aligned sequentially, where a given super-segment is used to give a starting point to the subsequent one. 
This is also shown in Fig.~\ref{fig:alignmentsquares}, where the second super-segment is aligned with the time-shift at the end of the first one. 
An additional benefit of segmentation is that it allows DCCA to scale proportionally to the input signal, thus linearly with the number of super-segments ($\mathcal{O}(Z)$).

In the following parts of the article we will detail how alignment is performed from the perspective of a single super-segment to keep index variables in check.

\subsection{Alignment}\label{sec:alignment}
The task of the alignment function is to give us estimates of $d(\tau_j) ~ \forall j \in [1,M]$, thus estimates of the local shift at any given knot point $\tau_j$. 
In our proposed alignment approach, we use the least-squares formulation from~\eqref{eq:NCC_CCA} and replace the projection matrix $W_i$ by the non linear function $f(\cdot, \theta_i)$ as was done with DCTW in~\eqref{eq:CTW}.
We proceed by minimizing the result for all signal segments $S_1^j, S_2^j$ for $j = 
1, \cdots M$
\begin{equation}
    \begin{array}{rl}
         \argmin\limits_{\theta, c^j} & \|f_1(S_1^j; \theta_1)\Delta_1^j - f_2(S_2^j; \theta_2)\Delta_2^j\|_F^2 \\
         \textrm{subject to} &\Delta_1^j \in \mathcal{D}_1^j(c),~~ \forall j \\
         &\Delta_2^j \in \mathcal{D}_2^j(c), ~~\forall j \\
    \end{array} \label{eq:NCC_applied}
\end{equation}

The transformation functions $f_i : \mathbb{R}^{l_i\times w} \rightarrow \mathbb{R}^{l_i \times w}, i\in \{1,2\} $ are represented by two neural networks with parameters $\theta_i$ and allow for the filtering of the input signals, and transform morphologically distinct signals to a correlated form. 
This is shown in part 2 of Fig.~\ref{fig:schematic}.
Finally, based on the NCC in~\eqref{eq:NCC_vec},
\begin{equation}
    \Sigma_{mn}^j = f_m(S_m^j; \theta_m)\Delta_m^j (\Delta_n^j)^T(f_n(S_n^j; \theta_n))^T
\label{eq:sigma_mn}
\end{equation}
this equation can be written as 
\begin{equation}
    NCC^j = \|(\Sigma_{11}^j)^{-1/2}\Sigma_{12}^j(\Sigma_{22}^j)^{-1/2}\|_{*}. \label{eq:DCCA}
\end{equation}
As $\Sigma_{mn}^j$ is a positive semidefinite matrix, the square root of the inverse matrix is calculated using the Cholesky decomposition. 
To ensure numerical stability, we furthermore incorporate a Tikhonov regularization if necessary.
By applying the derivations of~\cite[eq.(9)-(10)]{trigeorgis_deep_2015}, we can define an objective that allows for the optimal $\theta_{1,2}$ to be found which parameterize $f_{1,2}$. 
This leads to our optimized objective function 
\begin{equation}
    \begin{array}{rl}
         \max\limits_{\theta, c} \mathcal{L}_{NCC^j} &= \max\limits_{\theta, c} NCC^j,\\
         &= \textrm{trace}(NCC^j).
    \end{array} \label{eq:objective_raw}
\end{equation}
This optimization problem can now be solved with a neural network by computing the derivative of the objective function as
\begin{equation}
    \nabla \mathcal{L}_{NCC} = UV^{T}\textrm{trace}(\Sigma)
    \label{eq:derivative}
\end{equation}
where
\begin{equation}
     U, \Sigma, V = \textrm{SVD}\left((\Sigma_{11})^{-1/2}\Sigma_{12}(\Sigma_{22})^{-1/2}\right).
    \label{eq:svd_moment}
\end{equation}
The identity for the trace in $\eqref{eq:derivative}$ follows from the derivations listed in \cite{watson1992characterization}.
To train our neural network, and thus to find optimal transformation functions $f_1$ and $f_2$, we use backpropagation by solving for \eqref{eq:derivative}.
Note that in practice, while all possible NCC shifts are calculated simultaneously, we only take the derivative of the one corresponding to the maximal correlation. 

While optimizing the two networks simultaneously would be possible, it often becomes unstable and may result in overfitting behavior. 
Instead, we iteratively fix one network while learning the other and vice versa - with training on each side usually performed until convergence.
This optimization method leads to more stable results and in practice often requires at most two iterations but may be continued until convergence. 
Additionally, this iterative approach allows for segments with very low correlation values at each step to be filtered out.
This is desirable if it is expected that the two signals are only inter-correlated at some points in time.

We treat the choice of network architecture as a hyper-parameter. However, there are two recommendations: first, for efficiency reasons, it is advised to use a variant of Convolutional Neural Network (CNN); second, the receptive field of such a network should be locally constrained and thus no dilations or other ways of expanding it should be employed. In general, the network should only be able to process local parts of the signal to avoid introducing artificial signal shifts.

An important consideration when it comes to alignment is the use of linear correlations within short segments.
As a result, alignment performance can degrade heavily if a signal is non-stationary within single segments.
Luckily, in the vast majority of cases this can easily be addressed by first order differencing the non-stationary signal - in extreme cases one might even consider second order differencing.

\subsection{Constraining Alignment Variability}
One major problem we observed with the approach in~\ref{sec:alignment} is related to the flexibility of the neural networks $f_{1,2}$, which can lead to small internal time shifts between input and transformed output in terms of signal characteristics. 
This is comparable to one-sided digital filters that lead to a small signal shift. 
Although this behavior, as mentioned in~\ref{sec:alignment}, can be reduced by restricting the receptive field of a CNN, it is still observable, even if only within the chosen kernel size. 
To address this problem, we introduce a reconstruction penalty, the area loss $\mathcal{L}_{Area} = MSE(S_1, f_1(S_1, \theta_1))$.
This objective is only calculated on the signal whose network is not fixed, hence the asymmetry.
In cases where the dimensionality of network input and output do not match, an additional decoder can be used on the output before calculating the area loss.

\begin{figure}[!t]
\subfloat[]{\includegraphics[width=0.49\columnwidth]{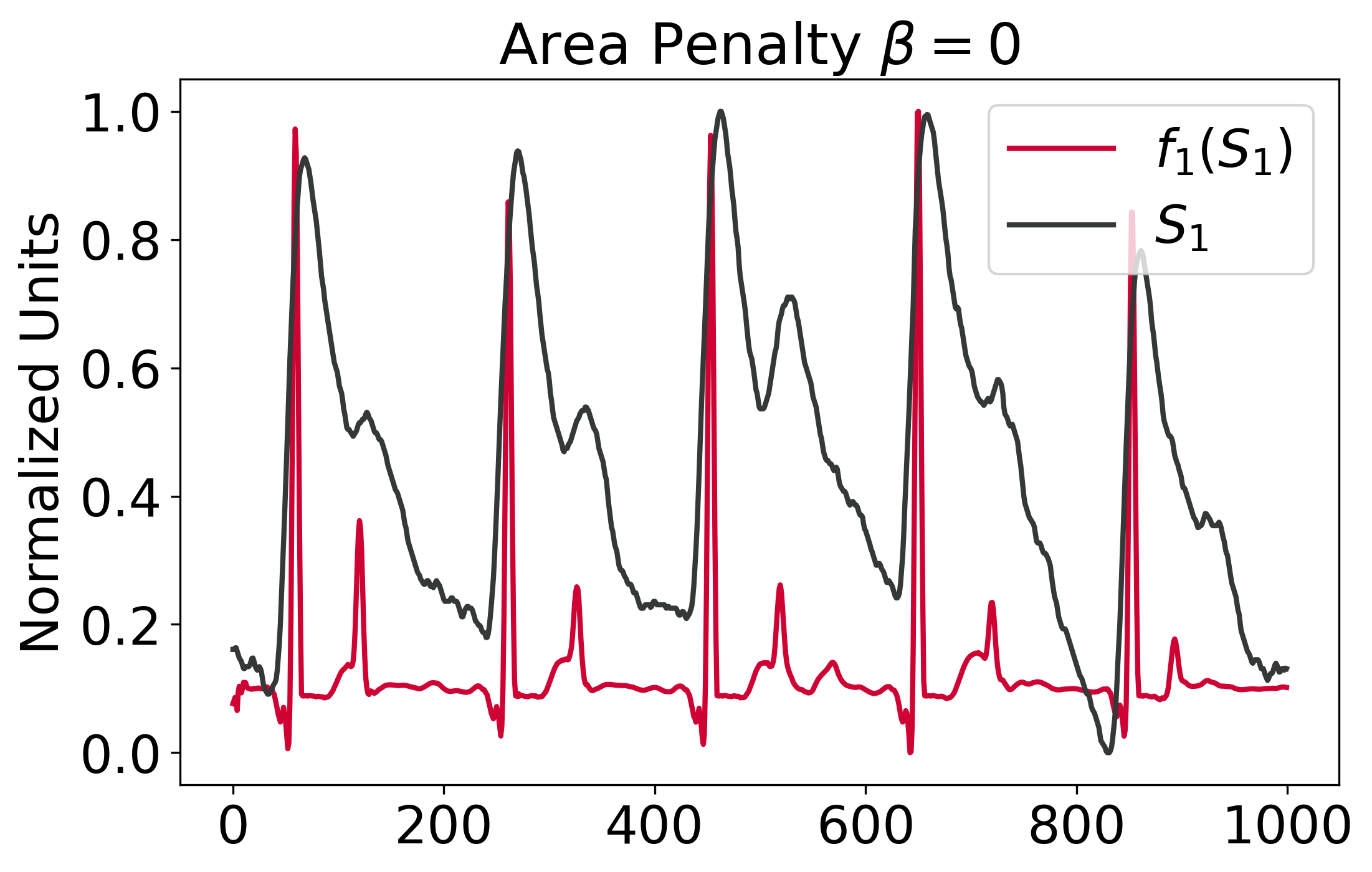}%
\label{fig:AreaLossPenalty}}
\hfil
\subfloat[]{\includegraphics[width=0.49\columnwidth]{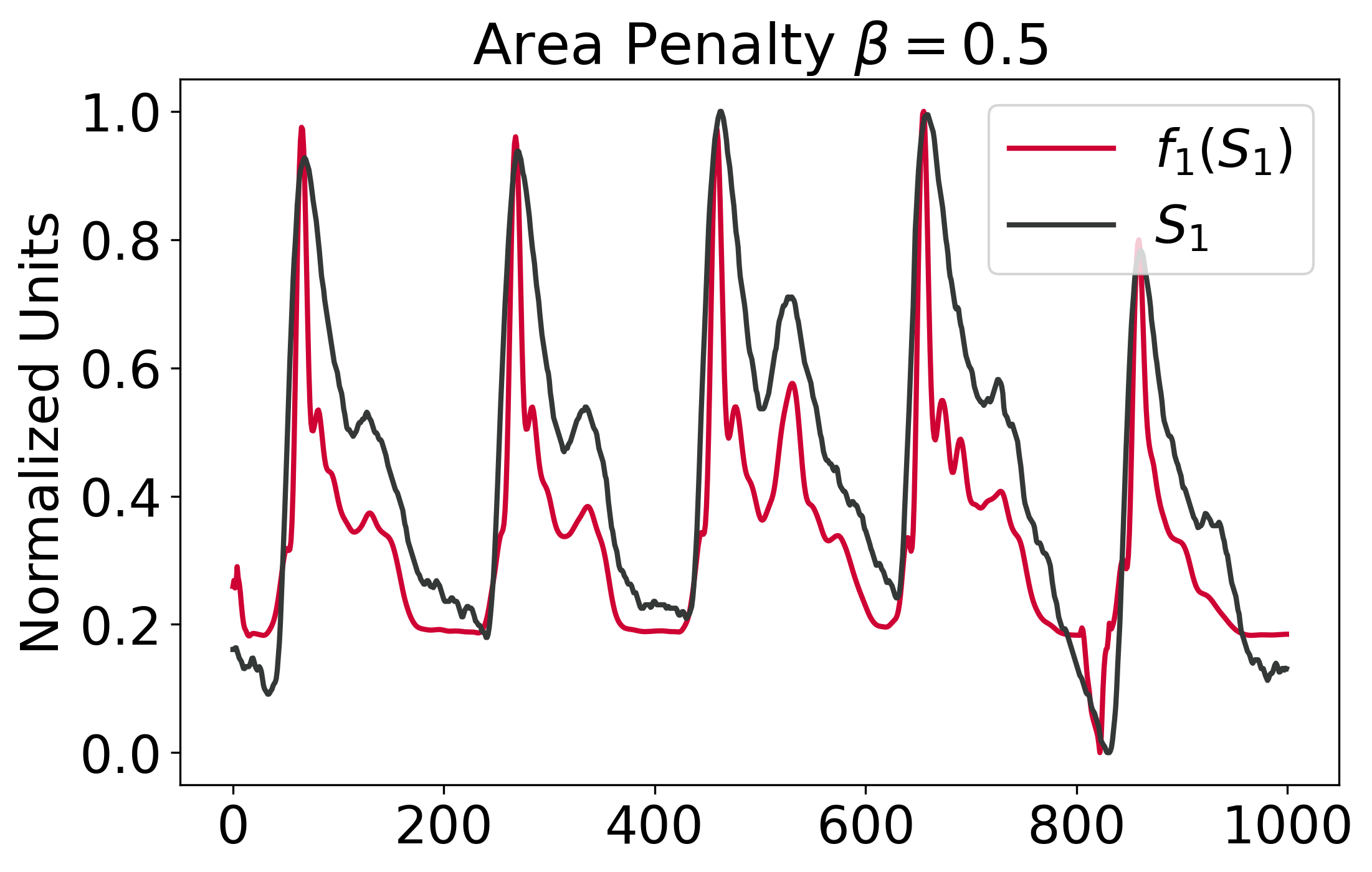}%
\label{fig:AreaLossBeta0}}
\hfil
\subfloat[]{\includegraphics[width=0.49\columnwidth]{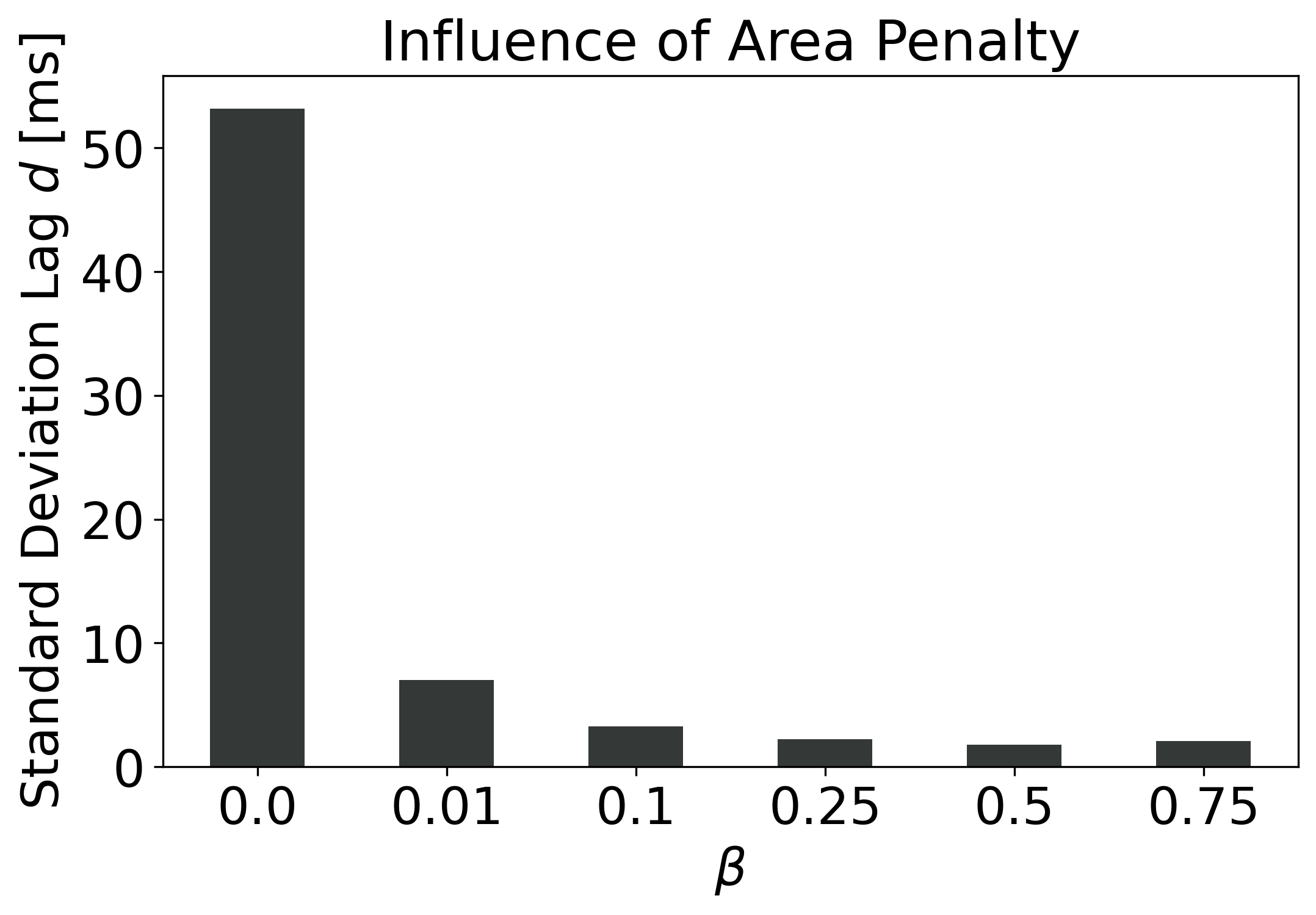}%
\label{fig:AreaLossBeta05}}
\subfloat[]{\includegraphics[width=0.49\columnwidth]{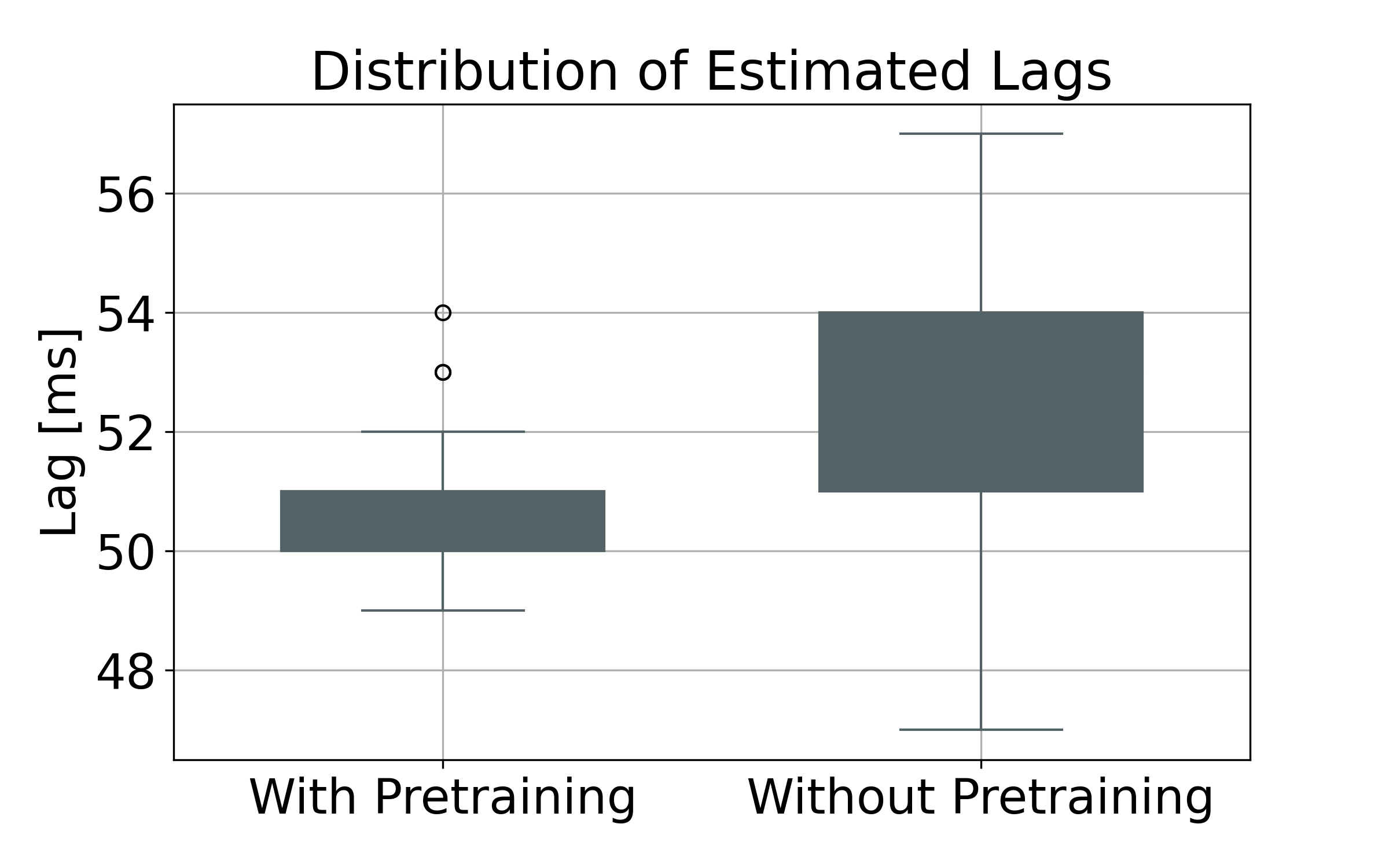}%
\label{fig:PretrainingLag}}
\caption{
a) Gives an example output of a transformation function trained with $\beta = 0$, while b) Shows the same with $\beta = 0.5$. One can see how in the unconstrained case with $\beta = 0$ the network $f_{1,2}$ morphs the PPG signal into a perfect ECG. However, when setting $\beta = 0.5$ it preserves characteristics of the input PPG signal as well as the target ECG signal. c) Shows the effect of the area penalty on alignment variability after repeating an ECG-PPG alignment 100 times for each choice of weighting parameter $\beta$ = {0, 0.01, 0.1, 0.25. 0.5, 0.75}. d) Highlights the distribution of lags of using a pre-trained $f_.$ (from another recording) compared to training from scratch (on the basis of two ECG-PPG recordings). Again, both conditions were repeated 100 times.}
\label{fig:areapenalty}
\end{figure}

Our final loss function is thus a convex combination of our proposed NCC and area loss. 
Therefore, for every window $j$ the objective $\mathcal{L}$ is optimized as

\begin{equation}
    \mathcal{L} = (1-\beta) \mathcal{L}_{NCC} + \beta \mathcal{L}_{Area}
    \label{eq:objective_full}
\end{equation}

where $\beta \in (0,1)$ is the weighting parameter.
Note that in the special case, setting $\beta = 1$, the objective \eqref{eq:objective_full} becomes a regular NCC (in the univariate case), as it forces $f_{1,2}$ to learn their identity functions.
The effect of the area penalty can be seen in Fig.~\ref{fig:areapenalty}.

Besides small internal shifts by a neural network, another source of alignment variability comes into play if two almost periodic signals are largely morphologically distinct. 
In this case, it is not clear which part of a cycle (such as a heart beat) from  $S_1$ should be matched with which part of the same cycle within $S_2$. 
Due to the stochasticity involved with neural network training, this can lead to different constant offsets over multiple applications of DCCA.

One way to address this behavior is to use a pre-trained model when aligning a new recording, which is often referred to as transfer learning.
To clarify, let's assume we have recordings from the same sensor types but from two different subjects and to each we apply DCCA for alignment.
Pre-training in this case means that we use the weights of the trained models from aligning recording \textit{1} as initial weights for the networks used to align record \textit{2}.
Fig.~\ref{fig:areapenalty} shows how using a pre-trained model reduces variability of the found alignment lag.

\subsection{Alignment Model Extraction}\label{sec:misalign}
As a result of~\ref{sec:alignment}, we are left with one estimate of warping function $d(\tau)$ at each knot point, thus we get a vector of estimates $\mathbf{\Delta} = [\hat{d}(\tau_1), \cdots, \hat{d}(\tau_M)]$.
Due to noise in the signals or the signals being only partially correlated, many entries in $\mathbf{\Delta}$ will not correspond to the actual signal misalignment. 
Directly correcting for the found misalignment would thus lead to many erroneous points in $\hat{d}(\cdot)$. 
However, we know that sensor clock-drift will likely stay similar to how it was before and after a noisy or uncorrelated interval.
Using this assumption we can search for a set of locally consistent functions of some family (such as linear or lower order polynomial) within $\mathbf{\Delta}$.
One can see this as a robust multi-model fitting problem, with, potentially very high, noise contamination. 
Our aim as such is to find a set of models $\mathcal{M}$ of some class that best describe the point estimates $\hat{d}(\tau)$ while disregarding noise.
Finding $\mathcal{M}$ is non-trivial.
Classic solutions like sequential RANSAC are greedy and often do not lead to satisfactory results~\cite{schutz2020real, isack_energy-based_2012}.

We therefore incorporate a non-greedy approach, using a global energy based multi-model fitting strategy, initially termed PE{\footnotesize A}RL~\cite{isack_energy-based_2012}.
The global energy function can be formulated as
\begin{equation}
    E(l) = \sum_{p \in \mathcal{P}} D_p(l_p) +\hspace{-0.1cm}\sum_{\{p,q\} \in \mathcal{N}}V_{\{p,q\}}(l_p,l_q) + h_L \cdot |\mathcal{M}| \label{eq:energy}
\end{equation}

\begin{figure}
\centering
\includegraphics[width=3in]{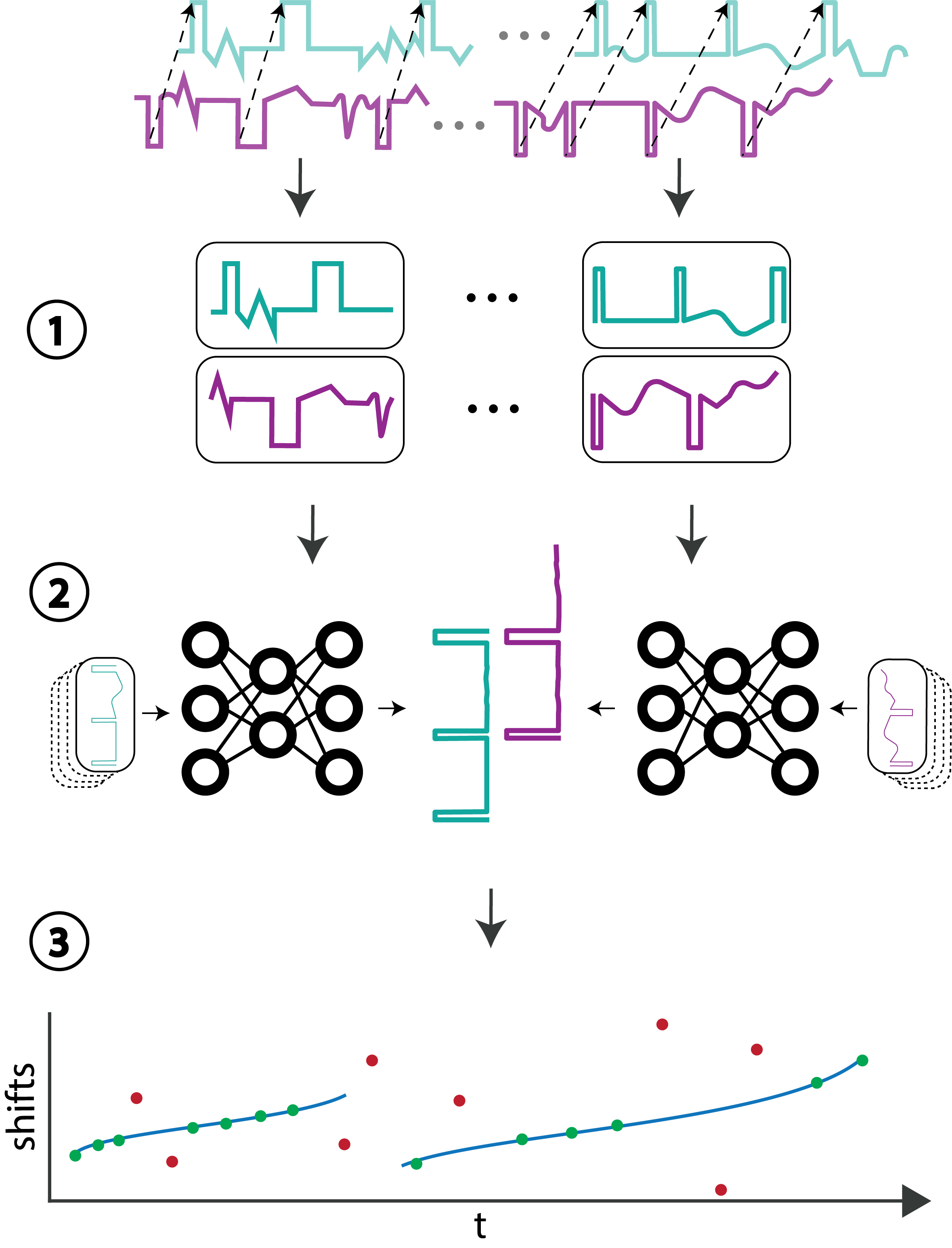}%
\caption{Displays a high-level overview of the proposed DCCA alignment algorithm. 
In (1), signals are joined on the time-axis of one sensor and segmented into smaller segments.
In (2) based on the proposed extended NCC objective two transformation networks are trained, calculating the shift between the two signals in a given segment as a by-product, while transforming both signals into a highly correlated signal space. 
This allows for the automatic filtering and transformation of signals to remove noise or account for morphological differences.
In (3), based on the resulting segment shifts, a noise tolerant energy based multi-model fitting approach is used to extract individual correction models which can be applied to the time-axis of the other sensor.}
\label{fig:schematic}
\end{figure}

In the case of fitting models, $D_p(l_p) = ||p-L||$ is the data cost, which measures how well a proposed model $L$ fits to a given data point $p$. 
$V_{\{p,q\}} = w_{pq}~\delta(L_p \neq L_q)$ is a smooth cost, that pushes adjacent data points in the neighbourhood $\mathcal{N}$ towards the same model (ensuring temporal consistency in our case), where $w_{pq} = exp\left(- \dfrac{||p-q||^2}{c^2}\right)$ accounts for the distance in $\mathcal{N}$, where $c$ is a weighting hyper-parameter for the smooth cost term. 
In our alignment case, $\mathcal{N}$ is a symmetric $n$-nearest neighbour graph on the basis of time points $t$. 
Finally, the last term $h_L \cdot |\mathcal{M}|$ penalizes the number of models in the final solution, where $h_L$ is the cost of a single model and $|\mathcal{M}|$ is the total number of selected models. 
Outliers are assigned to a noise model where $D_p$ is set to a constant value $\gamma$ that can intuitively be seen as the maximum allowed residual value for inlier-models.\newline
Eventually, $E(l)$ can be optimized with guaranteed optimality bounds using a modified $\alpha$-expansion algorithm \cite{delong2012fast, boykov2001fast}. 
As per the PE{\footnotesize A}RL algorithm, we iteratively switch between optimizing $E$ with $\alpha$-expansion and recomputing proposal models based on the previously found candidates. 
These two steps are repeated until convergence. 
This is part 3 in Fig.~\ref{fig:schematic}.
For more detailed explanations with regards to PE{\footnotesize A}RL, please refer to \cite{isack_energy-based_2012}. 
An important consideration here is that we can not only choose the family of functions, we want to fit, but also further restrict their hypothesis space - thus the infinite number of possible parameterizations.
As a result, this allows us to fit not only linear but also more complex functions like low order polynomials and further restrict, for instance, their maximal curvature.
In practice we found quadratic functions to work very well on real data, which matches observations by Vollmer et al. \cite{vollmer2019alignment}.

\section{Experiments}
To evaluate the performance of DCCA, we perform a number of experiments on both synthetic and real-world datasets.
We compare two versions of DCCA, one where two neural networks (one for $f_1$ and one for $f_2$) were trained, referred to as \textbf{B-DCCA}, and one where only $f_1$ was trained, while $f_2$ was fixed to the identity, referred to as \textbf{DCCA}. 
Additionally, to evaluate the usefulness of the transformation functions, we analyze the case where both $f_i$ are fixed to the identity, referred to as \textbf{I-DCCA}.\\
While we are not aware of a method made specifically for raw sensor signal alignment to benchmark DCCA against, we will still try to give the reader some idea of how other warping approaches compare.
We use two methods as baseline.
First, we take a piece-wise linear warping function approximation based on WCC, further referred to as \textbf{PLW}.
Second, we use a non-linear warping function approach based on an approximate DTW method \cite{salvador2007toward}, referred to as \textbf{NLW}.
Note that exact DTW methods would not be computationally feasible.
In cases with multivariate signals, \textbf{NLW} and \textbf{PLW} were extended with regular CCA \eqref{eq:CCA}.
To evaluate the accuracy, we calculated the mean absolute difference in milliseconds between the extracted approximation $\hat{d}[t]$ and the real misalignment $d[t]$, such that $MAE = \dfrac{1}{N}\sum_{t = 1}^{N}|d[t]-\hat{d}[t]|$.
Additionally, we report the standard deviation of the absolute differences. 
With multiple records, the grand-means are reported.
In the case of the DTW based \textbf{NLW}, this was based on the difference of the warping matrix (if multiple links collapsed to a single point, we used the mean difference) converted to milliseconds. 
For all experiments where stochasticity could play a role, each run was repeated five times. 
The transformation functions for DCCA and B-DCCA were approximated by a fairly standard 1D convolutional residual neural network, with one input layer, 15 intermediary layers (or blocks) as well as  three output convolutional layers.
We used the same hyperparameters across all experiments, except for window length $w$, which was slightly adjusted for each dataset.
The exact values used are provided in the supplementary material.

Additionally to the benchmark datasets, we show multiple examples of real-world alignments in Fig.~\ref{fig:SignalEvaluation}.

Furthermore, to give the reader some idea of real-world run-time of DCCA, we report the run-times of the alignment shown in Fig. \ref{fig:Signal4_raw} (training only $f_1$) on a machine with Ryzen 1950X CPU and Titan RTX GPU in Table \ref{table:benchmark}.

\subsection{Synthetic Datasets}
For the controlled evaluation of DCCA, we used synthetic datasets.\footnote{All synthetic datasets are available online~\cite{ARTSIG2021dataset} at https://zenodo.org/record/4522133.}
Synthetic datasets were meant to represent measurements obtained from observing the same event, but through different sensors or measurement technologies and under varying levels and types of noise. 
Additionally, clock-drift and offsets at random time points were added.
Different values for clock-drift and offsets were artificially added. 
Furthermore, the signals were distorted by three types of noises.
Moving baselines were added to make the signal non-stationary.
Segments of the signal were replaced by low-amplitude Gaussian noise to mimic signal-loss.
Segments of the signal were replaced by unrelated signals to simulate external noise factors.

\begin{figure}
\subfloat[]{\includegraphics[width=0.49\columnwidth]{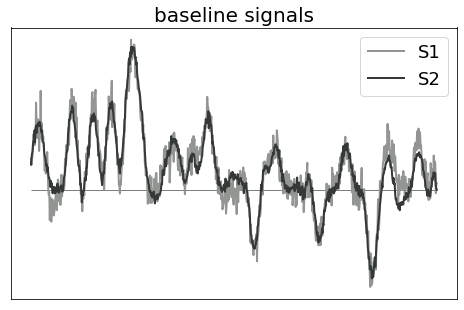}%
\label{fig:Baseline}}
\hfil
\subfloat[]{\includegraphics[width=0.49\columnwidth]{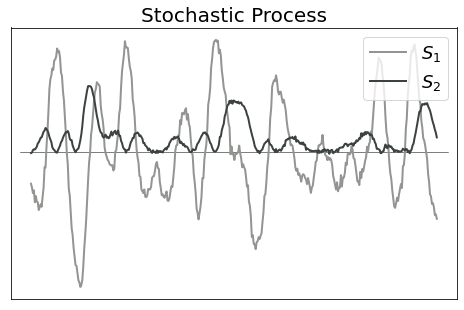}%
\label{fig:Poisson}}
\hfil
\subfloat[]{\includegraphics[width=0.49\columnwidth]{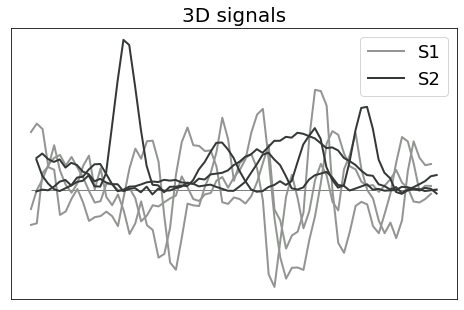}%
\label{fig:Art3D}}
\hfil
\subfloat[]{\includegraphics[width=0.49\columnwidth]{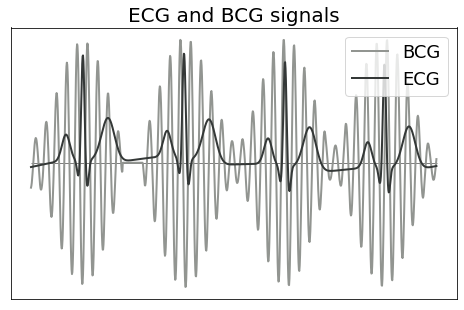}%
\label{fig:ArtECGBCG}}
\caption{Artificial signals used for algorithm testing. 
In \textbf{a)} is the baseline signal.
Part \textbf{b)} shows one dimensional Poisson processes.
In \textbf{c)}, 3D random movement signals are generated in Cartesian coordinates, $S_1$, and the derived spherical coordinates, $S_2$.
In \textbf{d)}, an artificial ECG is generated and a BCG signal is derived from it.}
\label{fig:ArtSignals}
\end{figure}

\paragraph{Baseline (SMP)} 
To prove, that all implementations are running correctly, a simple baseline dataset was created.
All algorithms should, by design, do well on this example.
The signals were created using multiple combined and differently weighted stochastic processes with Poisson distribution.
They were subsequently smoothed and  perturbed by the addition of Gaussian noise.
An example is given in Fig. \ref{fig:Baseline}.

\paragraph{Poisson-process 1D (Rnd1D)}
The basis of this test set is the same as the baseline set.
From the baseline, two signals were created, a difference signal and a squared signal.
A variety of noise and disturbing influences such as signal-loss, non-stationary-behaviour and external noise were mixed in.
See Fig. \ref{fig:Poisson}, for an example.

\paragraph{Poisson-process 3D (Rnd3D)}
The second test set is generated in the same way as the baseline signal, but with three dimensions for each signal, so $S_{1,2} \in \mathbb{R}^{3\times N}$.
Taking the first signal $S_1$ as the Cartesian coordinates of a movement pattern, the second signal $S_2$ was created by computing the spherical coordinates of the first signal.
To add further difficulties, the radius coordinate of the second signal was squared.
An example is given in Fig. \ref{fig:Art3D}.

\paragraph{Artificial ECG-BCG}
The third artificial dataset was inspired by real ECG-BCG measurements encountered in our own research~\cite{schutz2020real}.
The first signal is an artificially generated ECG signal, with variable heart-rate.
This signal was generated using ECGSYN~\cite{ECGSYN} based on work of~\cite{McSharry2003dynamical} and available on~\cite{PhysioNet}.
The second signal is an artificial BCG signal, derived from the first signal using wavelet-decomposition, smoothing, and sine-signal modulation.
An example is shown in Fig. \ref{fig:ArtECGBCG}.

\begin{table*}
\centering
  \caption{Alignment errors of different DCCA variations and a piece-wise linear (PLW) as well as non-linear (NLW) warping based baseline method on synthetic as well as real-world data.}
  \label{tab:results_artificial}
  \scriptsize
  \begin{tabular}{p{1.1cm}p{1.1cm}p{1.45cm}p{2.0cm}p{2.0cm}p{2.0cm}p{2.2cm}p{2.0cm}}
    \toprule
    Type & Morphology & Dataset & DCCA & B-DCCA & I-DCCA & PLW & NLW\\
    \midrule
    Synthetic & Similar & Baseline  &  17.3$\pm$135.1 ms & 85$\pm$134.0 ms  &  \textbf{15.0$\pm$135.2} ms & 426.1$\pm$2684.3 ms & 41.6$\pm$112.3 ms \\
    & Similar & Poisson 1D   &  11.5$\pm$116.1 ms & \textbf{11.0$\pm$116.5} ms &  122.1$\pm$298.2 ms & 3.9$\pm$6.7 s & 374.2$\pm$301.1 s \\
    & Similar & Poisson 3D &  \textbf{15.7$\pm$76.0} ms & 19.8$\pm$76.3 ms &  308.2$\pm$275.4 ms & 0.9$\pm$1.6 s & 33.8$\pm$24.0 s \\
    & Distinct & ECG-BCG  &  \textbf{228.0$\pm$114.1} ms &  329.2$\pm$140.6 ms &  7.8$\pm$4.2 s & 16.5$\pm$10.0 s & 18.7$\pm$12.9 s \\
    \midrule
    Real    & Similar & ECG-ECG 1  &  \textbf{3.8$\pm$1.6} ms& \textbf{3.8$\pm$1.6} ms & \textbf{3.8$\pm$1.6} ms & 259.8$\pm$861.1 ms & 194.9$\pm$135.7 s\\
    & Similar & ECG-ECG 2  &  19.8$\pm$2.1 ms & 49.6$\pm$2.5 ms & \textbf{2.8$\pm$1.6} ms & 1.2$\pm$1.8 s & 244.3$\pm$175.3 s\\
    & Distinct & PPG-ECG  &  \textbf{318.3$\pm$24.7} ms &  434.0$\pm$84.3 ms  &  10.3$\pm$2.0 s & 3.0$\pm$2.8 s & 192.7$\pm$93.6 s\\
    & Distinct & aVL-ECG  &  \textbf{50.7$\pm$2.1} ms & 96.2$\pm$2.3 ms &  2.3$\pm$1.4 s & 4.1$\pm$2.8 s & 108.0$\pm$70.9 s\\
    & Distinct & ECG-ACC\footnote{*}  &  \textbf{144.1$\pm$12.1} ms & 170.7$\pm$12.4 ms &  3.3$\pm$2.0 s & 8.8$\pm$9.3 s & 2.5$\pm$0.246 s\\
  \bottomrule
\end{tabular}
\end{table*}
\footnotetext{ the median across is reported instead of the mean due to large outliers across all methods}

\subsection{Real Datasets}
While we have evaluated DCCA on vast quantities of our own data, regulations prohibit those datasets from being shared. 
To still keep evaluation on real data as transparent as possible we resort to a publicly available dataset, where the authors share semi-manually aligned raw sensor signals together with the unaligned ones \cite{vollmer2019alignment, PhysioNet}.
The dataset covers recordings from 13 people (of which we use the last 9) doing a set of different tasks, including resting, walking, and cognitive tasks while being measured with a variety of sensors. 
Each recording is approximately half an hour long.
The authors provide semi-manual ground-truth alignments, based on R-R intervals. 
For our experiments, we calculated the misalignment $d(t)$ of the respective raw data and the ones the authors already aligned. 
As the authors note, some of the sensors do exhibit non-linear sampling frequencies and as such make for an ideal case to evaluate the proposed DCCA algorithm.
We constructed 5 benchmark scenarios.
We used the medical grade NeXus-10 MKII (Mind Media, Netherlands) ECG device, as a reference sensor, and aligned various signals of the eMotion Faros 360° (Mega, Finland) as well as the SOMNOtouch NIBP (Somnomedics, Germany) to it.
The first case (\textit{ECG-ECG 1}) consists of the ECG signals of the NeXus and the Faros device.
The second case \textit{ECG-ECG 2} is made with the ECG signals of the NeXus and SOMNOtouch device.
The third case (\textit{PPG-ECG}) requests alignment between the ECG of the NeXus and the PPG signal of the SOMNOtouch device.
The fourth case (\textit{aVL-ECG}) is made of the ECG of the NeXus and a morphologically distinct lead (aVL) of the SOMNOtouch device.
The fifth case (\textit{ECG-ACC}) involves the alignment of the ECG from the NeXus with the 3-Axis Accelerometer of the Faros device.
In the case of the PPG signal, we noticed sometimes very strong non-linear behavior, likely due to movement, we therefore applied
first order differencing in this scenario.
More detailed information about the datasets is provided in the supplementary material.

\begin{figure*}[!t]
\subfloat[]{\includegraphics[width=0.245\columnwidth]{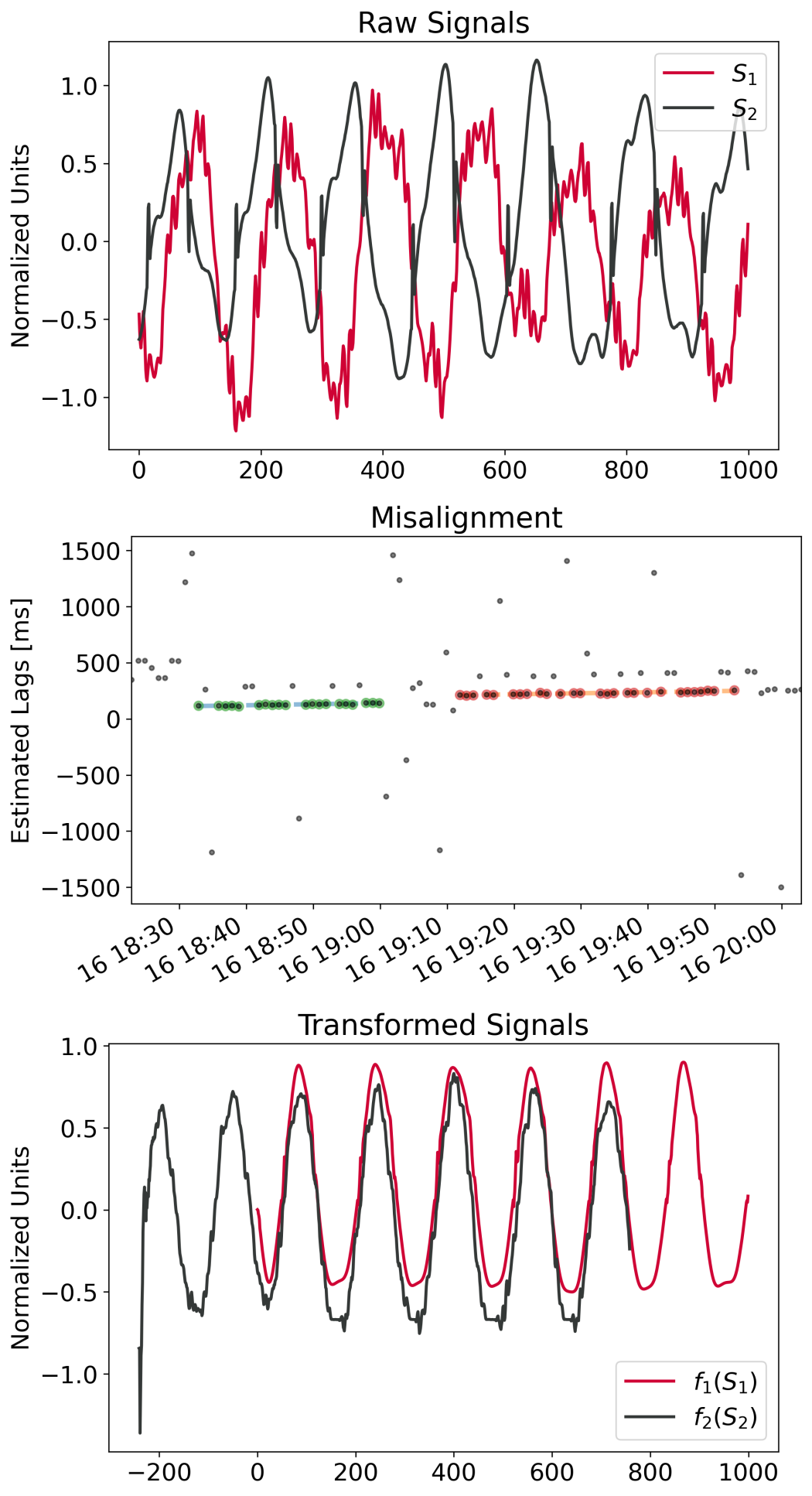}%
\label{fig:Signal1_raw}}
\hfil
\subfloat[]{\includegraphics[width=0.245\columnwidth]{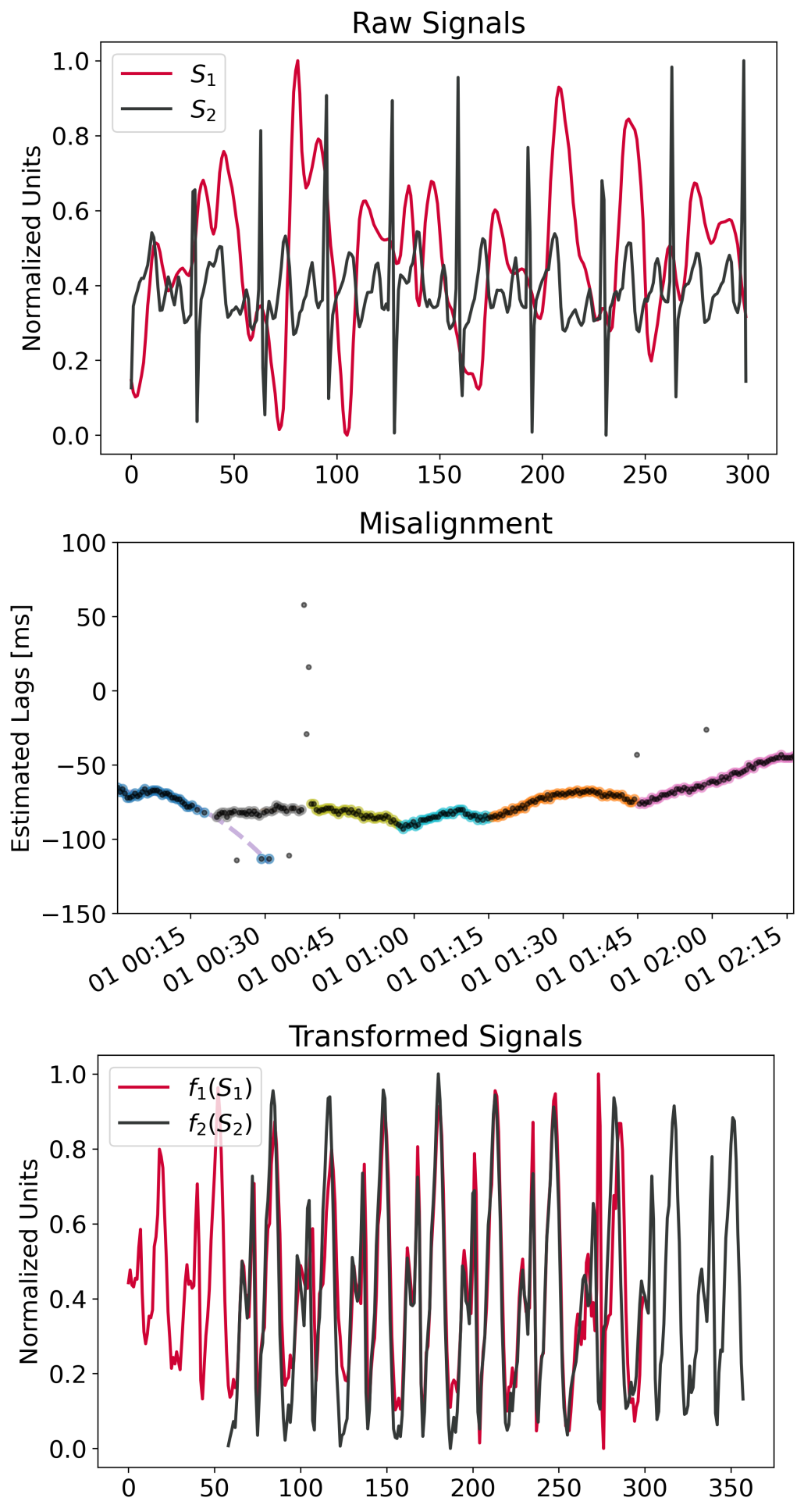}%
\label{fig:Signal2_raw}}
\hfil
\subfloat[]{\includegraphics[width=0.245\columnwidth]{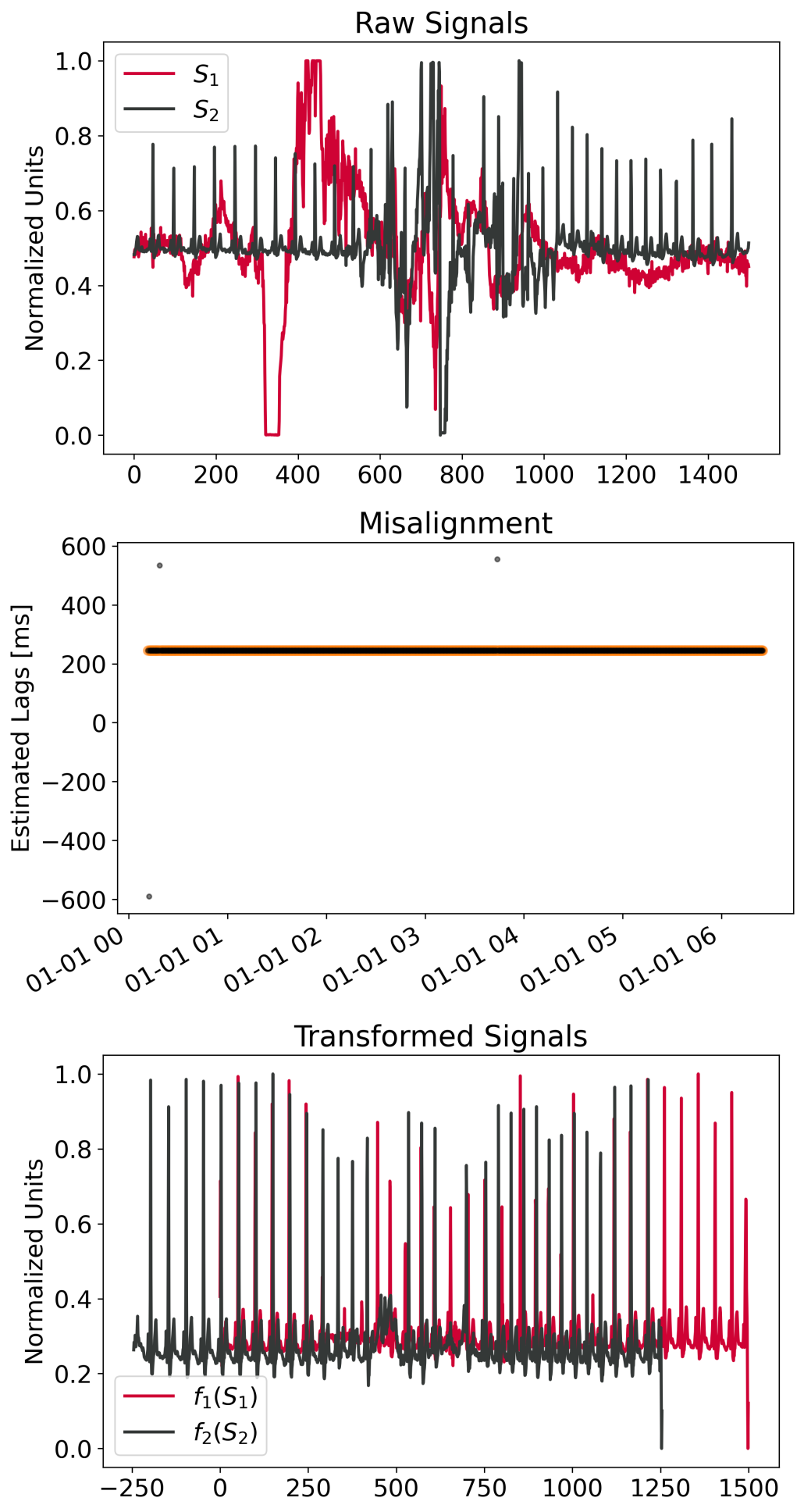}%
\label{fig:Signal3_raw}}
\hfil
\subfloat[]{\includegraphics[width=0.245\columnwidth]{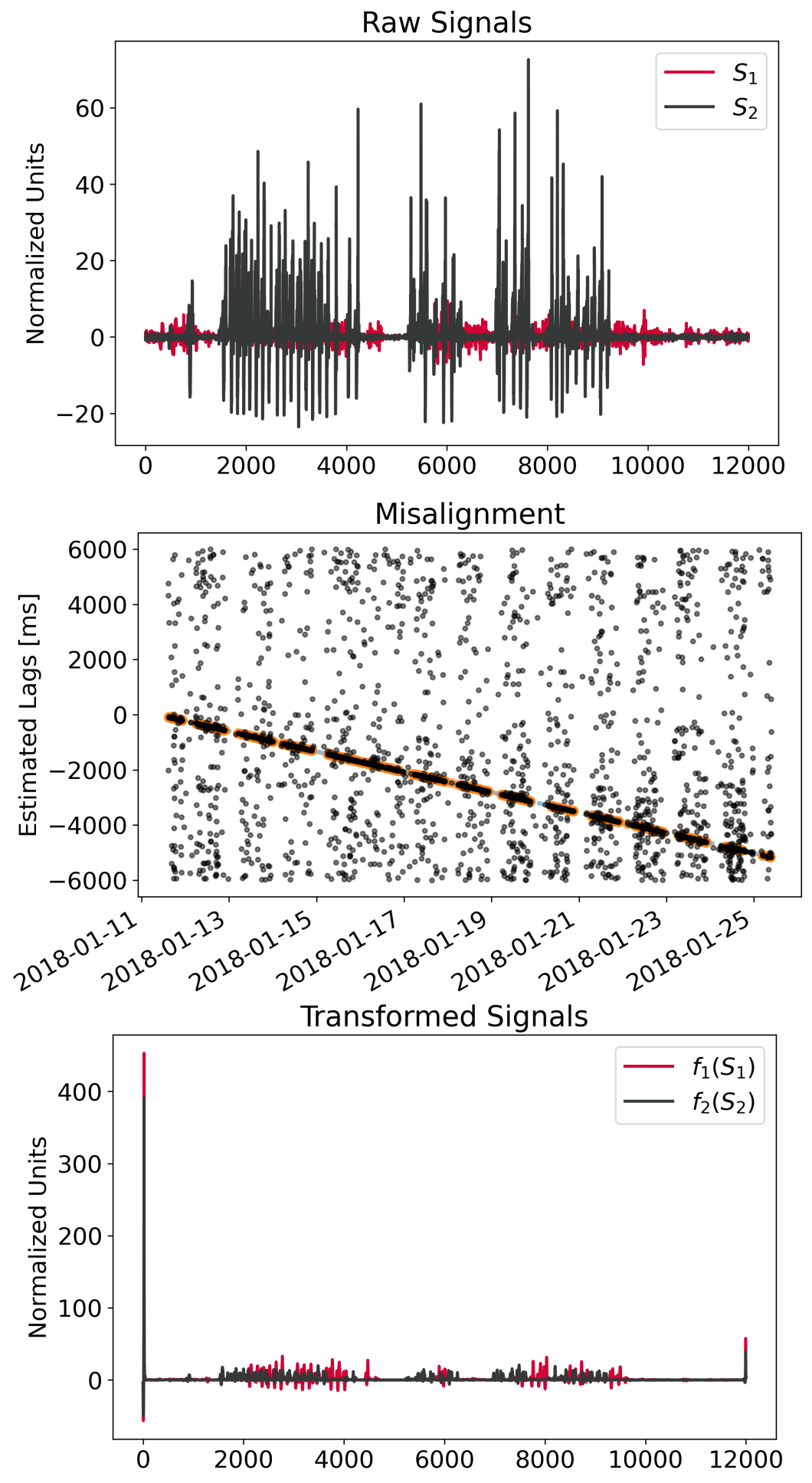}%
\label{fig:Signal4_raw}}
\caption{
In the first row, the raw signals are depicted. 
In the second row, the time-shifts $d$ for each window are depicted.
The alignment models are indicated by coloured lines, with the points and corresponding model in the same colour. Outliers are black.
In the third row, the signal outputs from the neural network are shown.
Due to the filtering and morphing properties of the network, the signals look more alike. They are shifted according to the computed time-shift $d$.
a) The signals in the first column represent an alignment between respiration measured with a contactless consumer grade BCG-device and a clinical grade monitoring device.
b) Shows an alignment between an ECG and a PPG sensing device, exhibiting non-linear clock-drift, obtained from \cite{PhysioNet}.
c) Highlights the alignment between an EEG and an ECG sensor, measured by the same polysomnography measurement device. Here an artificial offset was introduced, to give a negative example of an alignment without any clock-drift \cite{schmidt2018introducing}.
d) Here we depict the alignment between two accelerometers with linear clock-drift behavior, one worn on the ankle, the other on the wrist \cite{botros2019long}.
}
\label{fig:SignalEvaluation}
\end{figure*}

\begin{table}
\centering
  \caption{A listing of run-times of DCCA along different stages of the algorithm. Here two signals with 120'000'000 data points were aligned on a consumer grade machine.}
  \label{table:benchmark}
  \begin{tabular}{llll}
    \toprule
    Segmentation & Alignment & Model Extraction & Total\\
    \midrule  
    02:00 min & 19:36 min &  03:48 min & \underline{\textbf{25:24}} min\\
  \bottomrule
\end{tabular}
\end{table}

\subsection{Results and Discussion}
Results on both synthetic and real datasets are presented in Table~\ref{tab:results_artificial}.
It is apparent, that under ideal conditions, all methods can align the synthetically generated data, with the exception of \textbf{PLW}, all within 100 milliseconds of the true alignment.
In case of \textbf{PLW}, the worst result is due to large individual outliers.
When it comes to more difficult cases with severe noise, as well as with real-world signals, performance of \textbf{I-DCCA}, \textbf{PLW}, and \textbf{NLW} based alignments drops strongly. 
It is further clearly noticeable, how in the morphologically strongly distinct ECG-BCG, PPG-ECG, aVL-ECG and ECG-ACC datasets, the inclusion of transformation functions $f_i$ of DCCA based approaches lead to the lowest alignment errors by a large margin, when compared to DCCA with no transformations (\textbf{I-DCCA}).
In cases with very little morphological differences, such as with Baseline, ECG-ECG 1 and ECG-ECG 2, we found (\textbf{I-DCCA}) without trainable transformation functions to be slightly more accurate.
We suspect that this is because of the variability introduced as a result of neural network training.
While both \textbf{DCCA} and \textbf{B-DCCA} are competitive on all datasets, differences between them are not too big, with \textbf{B-DCCA} showcasing consistently higher errors - again, likely as a result of the additional alignment variability brought about by the two trainable $f_i$. 
It is important to remark that in the case of PPG and ACC signals, a perfect alignment is not possible as we calculated the true alignment based on the ECG signals (as this is what the original authors provide). 
Due to this, there is a variable time delay between the registration of the electrical ECG signal and the arrival time of the pulse wave that the optical/mechanical signals of the PPG and accelerometer represent. 
As such, alignments within a single heartbeat (approximately 1000ms) should be considered accurate for the PPG and accelerometer.
Additionally, all methods struggled with the ACC-ECG based alignment, although to varying degrees (note that we had to use the median due to too many heavy outliers, where no alignment was found).
This intuitively makes sense, as it shows an extremely difficult alignment scenario because the signals are only very weakly correlated and the used accelerometers were not configured to measure cardiac activity at all.
In this context, the performance of the ACC-ECG alignment is quite remarkable.
Nonetheless, this also shows the limitations of DCCA in general.
If there simply is no correlation or it is very weak, a sub-second alignment error may not be achievable.

Overall DCCA approaches perform an order of magnitude better compared to the baselines, showcasing sub-second alignment errors, which allows even finer signal characteristics like individual heart-beats to be matched.
When it comes to DCCA variations, it seems that for many cases a single transformation function is sufficient and even beneficial.
In cases with little signal noise and minimal morphological differences, setting both transformations to their identity may further lead to some smaller reductions in alignment errors.
The examples in Fig.~\ref{fig:SignalEvaluation}, further show how DCCA is applicable to other alignment scenarios in the biomedical field, leading to visually convincing alignments. 

\section{Proof of Concept Study - Self-Supervised Learning}
While aligning raw signals itself may already provide useful in validation scenarios, one of the most promising areas, that automated alignment methods like DCCA can enable, is self-supervised learning (SSL).
SSL is becoming increasingly relevant and heavily used in a variety of settings, such as language modelling and more recently also in computer vision. 
As such, Facebook AI researchers have called SSL a very promising candidate to unlock the "dark matter of intelligence"  \cite{lecun_misra}.
In contrast to how SSL is used in other domains, we here refer to a variant of SSL, where no human labeling is necessary but instead, the labels are based on a validated sensor signal itself. 
In the context of biomedical signals, we use a gold-standard, validated sensor to provide machine labeled data that may then be used to train models on another, independent, sensor.
This could allow for the creation of very large machine annotated datasets with very little human effort, apart from having subjects monitored by multiple sensors.
Note, the big advantage here is that we can use "ANY" existing sensor, without having to consider low-level synchronization.
To highlight this, we demonstrate DCCA on a simple R-peak extraction task.
Given raw data from a contactless BCG sensor and a validated hospital grade ECG device, the aim is to learn a model that extracts R-peaks from the BCG signal based on automatically labeled ECG R-peaks.
For this we used 2'178 minutes of simultaneous ECG-BCG recordings as training set and a separate 161 minutes as holdout for testing.
This dataset was obtained from people lying on a hospital bed, while being attached to a clinical grade ECG device and at the same time having a BCG sensor placed below the mattress.
More details for reproducibility are given in the supplementary material.

\begin{figure}[h!]
\centering
\includegraphics[width=0.5\columnwidth]{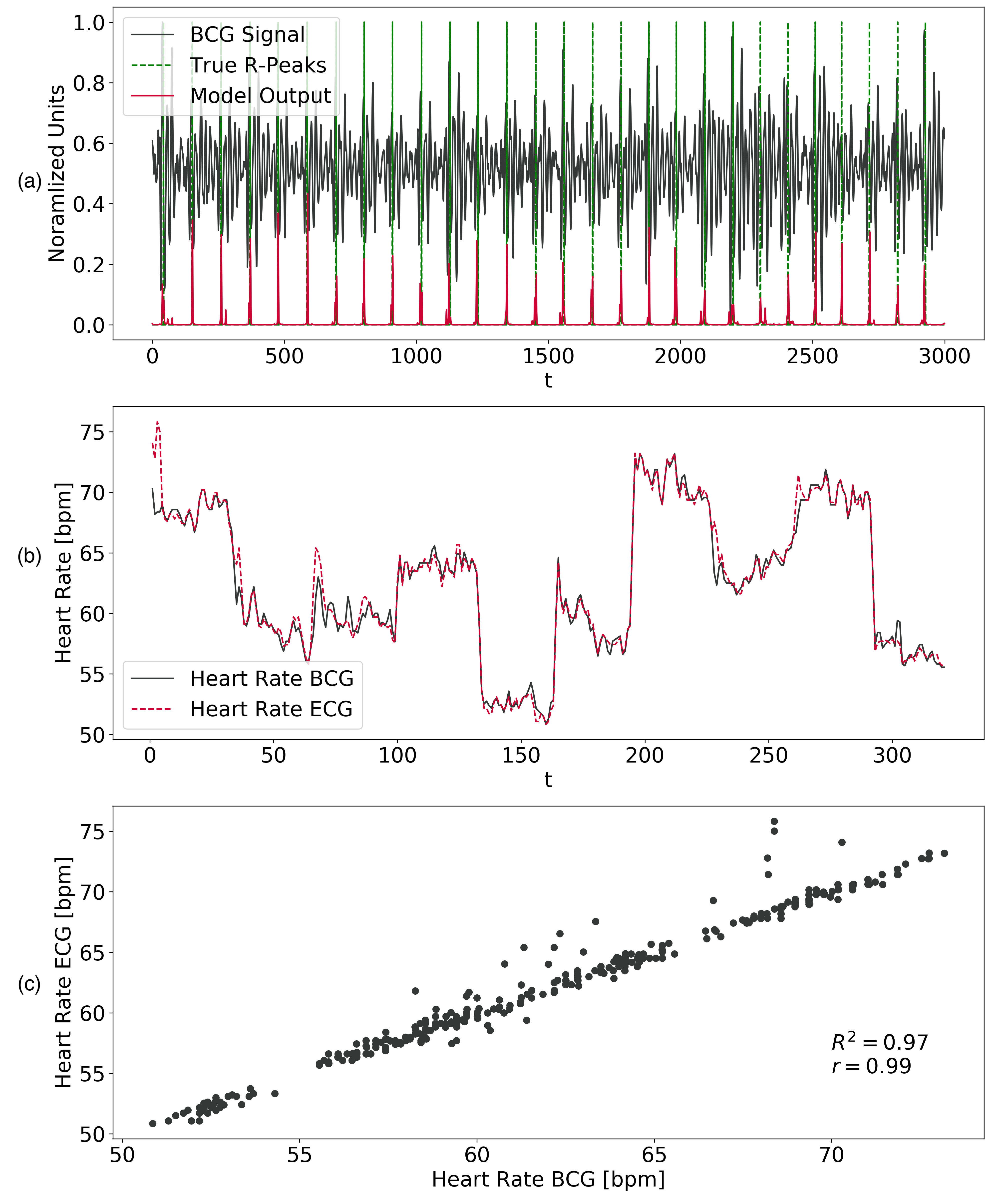}%
\caption{Depiction of results from the self-supervised R-peak extraction proof of concept study. a) shows an example of a 30 segment, containing the BCG input signal, together with the corresponding peak predictions from the learnt model as well as the ECG derived R-peaks that serve as labels. b) highlights a comparison between the extracted heart rate on the basis of the model predictions (BCG) and the same for the ECG derived R-peaks (both on the holdout set). c) shows the extracted BCG heart rates, plotted against the same from the ECG signal. Additionally, the coefficient of determination $R^2$ and Pearson's correlation coefficient $r$ between the two is shown.}
\label{fig:case_study}
\end{figure}

For this task, the data was first resampled to a common sampling frequency of 100 Hz and then aligned with DCCA (using the same default hyper parameters as for the rest of the article).
Next, a set of 30 second segments was extracted, where the raw BCG signal (with segment-wise robust scaling), represents the input and the ECG R-peaks the target labels.
For the learning task we employed a 1D U-Net-BiDirectional-LSTM model (for details see supplementary material).
Finally, to extract discrete peaks from the model output, we use a simple peak detector (for details please refer to the supplementary material).
Based on these peaks, the heart rate per segment is calculated.
Eventually to smooth out smaller perturbations, we apply a 1-minute moving median filter on both ECG and BCG derived heart rate.
Please note, that this is merely a proof of concept, model architecture, hyper parameters and post-processing could very likely be largely improved upon.
In Fig.~\ref{fig:case_study}, we show the result, on the holdout data.
As one can see, results are already surprisingly good, with an $R^2$ of 0.97 and a Pearson correlation coefficient of 0.99.
Additionally, the mean absolute error between the ECG and BCG derived heart rate is relatively low with 0.49 beats per minute (bpm).
As a sanity check, we applied the same procedure to the non-aligned raw data and got, as expected, R-peak extraction models that did not converge at all.
While this is an informal example, with more elaborate algorithms and potentially much more data, one could very well imagine using a similar approach to extract highly relevant information, for example heart arrhythmias, in a fully automated fashion.

\section{Conclusion}
In this work we introduce the first (to the best of our knowledge) algorithm specifically designed to correct for temporal misalignments between inter-correlated, raw sensor signals. 
The proposed approach has the ability to automatically filter and transform signals in a non-linear fashion, such that the correlation between them is maximized. 
Furthermore, it is fast enough to handle several hundred million data points per hour on decent modern hardware.
As a result, even highly noisy and morphologically distinct signals may be accurately corrected for internal clock-drift and offsets. 
We showcase the applicability of the proposed solution on multiple synthetic and real world datasets, where average alignment error was in the sub-second resolution or even sub 100 millisecond one, outperforming baseline approaches by an order of magnitude.
Furthermore, applicability of DCCA is demonstrated on additional real-world alignment scenarios showcasing convincing visual results.
Lastly, we show the usability and potential of DCCA in a self-supervised learning scenario where alignment performance is accurate enough to correctly learn an R-peak extractor on morphologically distinct BCG signals, in a fully automated fashion - without need for human labeling or alignment.

\section*{Acknowledgment}
Thanks to all other members of our team, and especially Samuel Knobel for the valuable help.

\newpage

\bibliographystyle{IEEEtran}
\bibliography{DCCSA_bibliography.bib}

\begin{thebibliography}{10}
\providecommand{\url}[1]{#1}
\csname url@samestyle\endcsname
\providecommand{\newblock}{\relax}
\providecommand{\bibinfo}[2]{#2}
\providecommand{\BIBentrySTDinterwordspacing}{\spaceskip=0pt\relax}
\providecommand{\BIBentryALTinterwordstretchfactor}{4}
\providecommand{\BIBentryALTinterwordspacing}{\spaceskip=\fontdimen2\font plus
\BIBentryALTinterwordstretchfactor\fontdimen3\font minus
  \fontdimen4\font\relax}
\providecommand{\BIBforeignlanguage}[2]{{%
\expandafter\ifx\csname l@#1\endcsname\relax
\typeout{** WARNING: IEEEtran.bst: No hyphenation pattern has been}%
\typeout{** loaded for the language `#1'. Using the pattern for}%
\typeout{** the default language instead.}%
\else
\language=\csname l@#1\endcsname
\fi
#2}}
\providecommand{\BIBdecl}{\relax}
\BIBdecl

\bibitem{ahmed2017role}
E.~Ahmed, I.~Yaqoob, I.~A.~T. Hashem, I.~Khan, A.~I.~A. Ahmed, M.~Imran, and
  A.~V. Vasilakos, ``The role of big data analytics in internet of things,''
  \emph{Computer Networks}, vol. 129, pp. 459--471, December 2017.

\bibitem{rantz2015new}
M.~J. Rantz, M.~Skubic, M.~Popescu, C.~Galambos, R.~J. Koopman, G.~L.
  Alexander, L.~J. Phillips, K.~Musterman, J.~Back, and S.~J. Miller, ``A new
  paradigm of technology-enabled ‘vital signs' for early detection of health
  change for older adults,'' \emph{Gerontology}, vol.~61, no.~3, pp. 281--290,
  2015.

\bibitem{saner2021case}
H.~Saner, N.~Schuetz, P.~Buluschek, G.~Du~Pasquier, G.~Ribaudo, P.~Urwyler, and
  T.~Nef, ``Case report: Ambient sensor signals as digital biomarkers for early
  signs of heart failure decompensation,'' \emph{Frontiers in cardiovascular
  medicine}, vol.~8, p.~11, 2021.

\bibitem{schutz2021contactless}
N.~Sch{\"u}tz, H.~Saner, A.~Botros, B.~Pais, V.~Santschi, P.~Buluschek,
  D.~Gatica-Perez, P.~Urwyler, R.~M. M{\"u}ri, T.~Nef \emph{et~al.},
  ``Contactless sleep monitoring for early detection of health deteriorations
  in community-dwelling older adults: Exploratory study,'' \emph{JMIR mHealth
  and uHealth}, vol.~9, no.~6, p. e24666, 2021.

\bibitem{evidation2019}
R.~Chen, F.~Jankovic, N.~Marinsek, L.~Foschini, L.~Kourtis, A.~Signorini,
  M.~Pugh, J.~Shen, R.~Yaari, V.~Maljkovic \emph{et~al.}, ``Developing measures
  of cognitive impairment in the real world from consumer-grade multimodal
  sensor streams,'' in \emph{Proceedings of the 25th ACM SIGKDD International
  Conference on Knowledge Discovery \& Data Mining}, 2019, pp. 2145--2155.

\bibitem{amin2016radar}
M.~G. Amin, Y.~D. Zhang, F.~Ahmad, and K.~D. Ho, ``Radar signal processing for
  elderly fall detection: The future for in-home monitoring,'' \emph{IEEE
  Signal Processing Magazine}, vol.~33, no.~2, pp. 71--80, 2016.

\bibitem{boehme2019soon}
P.~Boehme, A.~Hansen, R.~Roubenoff, J.~Scheeren, M.~Herrmann, T.~Mondritzki,
  J.~Ehlers, and H.~Truebel, ``How soon will digital endpoints become a
  cornerstone for future drug development?'' \emph{Drug discovery today},
  vol.~24, no.~1, pp. 16--19, 2019.

\bibitem{coravos2020modernizing}
A.~Coravos, M.~Doerr, J.~Goldsack, C.~Manta, M.~Shervey, B.~Woods, and W.~A.
  Wood, ``Modernizing and designing evaluation frameworks for connected sensor
  technologies in medicine,'' \emph{NPJ digital medicine}, vol.~3, no.~1, pp.
  1--10, 2020.

\bibitem{henriksen2018using}
A.~Henriksen, M.~H. Mikalsen, A.~Z. Woldaregay, M.~Muzny, G.~Hartvigsen, L.~A.
  Hopstock, and S.~Grimsgaard, ``Using fitness trackers and smartwatches to
  measure physical activity in research: analysis of consumer wrist-worn
  wearables,'' \emph{Journal of medical Internet research}, vol.~20, no.~3, p.
  e9157, 2018.

\bibitem{king2018survey}
C.~E. King and M.~Sarrafzadeh, ``A survey of smartwatches in remote health
  monitoring,'' \emph{Journal of healthcare informatics research}, vol.~2,
  no.~1, pp. 1--24, 2018.

\bibitem{shin2019wearable}
G.~Shin, M.~H. Jarrahi, Y.~Fei, A.~Karami, N.~Gafinowitz, A.~Byun, and X.~Lu,
  ``Wearable activity trackers, accuracy, adoption, acceptance and health
  impact: A systematic literature review,'' \emph{Journal of biomedical
  informatics}, vol.~93, p. 103153, 2019.

\bibitem{bent2020investigating}
B.~Bent, B.~A. Goldstein, W.~A. Kibbe, and J.~P. Dunn, ``Investigating sources
  of inaccuracy in wearable optical heart rate sensors,'' \emph{NPJ Digital
  Medicine}, vol.~3, no.~1, pp. 1--9, 2020.

\bibitem{jiang2020eventdtw}
Y.~Jiang, Y.~Qi, W.~K. Wang, B.~Bent, R.~Avram, J.~Olgin, and J.~Dunn,
  ``Eventdtw: An improved dynamic time warping algorithm for aligning
  biomedical signals of nonuniform sampling frequencies,'' \emph{Sensors},
  vol.~20, no.~9, p.~13, May 2020.

\bibitem{vollmer2019alignment}
M.~Vollmer, D.~Bl{\"a}sing, and L.~Kaderali, ``Alignment of multi-sensored
  data: adjustment of sampling frequencies and time shifts,'' in \emph{2019
  Computing in Cardiology (CinC)}.\hskip 1em plus 0.5em minus 0.4em\relax
  Singapore, Singapore: IEEE, 2019, pp. Page--1.

\bibitem{schutz2020real}
N.~Sch{\"u}tz, A.~A. Botros, S.~E. Knobel, H.~Saner, P.~Buluschek, and T.~Nef,
  ``Real-world consumer-grade sensor signal alignment procedure applied to
  high-noise ecg to bcg signal synchronization,'' in \emph{2020 42nd Annual
  International Conference of the IEEE Engineering in Medicine \& Biology
  Society (EMBC)}.\hskip 1em plus 0.5em minus 0.4em\relax Montreal, Canada:
  IEEE, August 2020, pp. 5858--5962.

\bibitem{zhou2008frequency}
H.~Zhou, C.~Nicholls, T.~Kunz, and H.~Schwartz, ``Frequency accuracy \&
  stability dependencies of crystal oscillators,'' Carleton University,
  Ottawa,Canada, Systems and Computer Engineering Technical Report SCE-08-12,
  2008.

\bibitem{sivrikaya2004time}
F.~Sivrikaya and B.~Yener, ``Time synchronization in sensor networks: a
  survey,'' \emph{IEEE network}, vol.~18, no.~4, pp. 45--50, 2004.

\bibitem{lasassmeh2010time}
S.~M. Lasassmeh and J.~M. Conrad, ``Time synchronization in wireless sensor
  networks: A survey,'' in \emph{Proceedings of the IEEE SoutheastCon 2010
  (SoutheastCon)}, IEEE.\hskip 1em plus 0.5em minus 0.4em\relax Concord, NC,
  USA: IEEE, 2010, pp. 242--245.

\bibitem{harke1999cardiac}
K.~Harke, A.~Schl{\"o}gl, P.~Anderer, and G.~Pfurtscheller, ``Cardiac field
  artifact in sleep eeg,'' \emph{Proceedings EMBEC’99}, pp. 482--483, 1999.

\bibitem{huysmans2019evaluation}
D.~e.~a. Huysmans, ``Evaluation of a commercial ballistocardiography sensor for
  sleep apnea screening and sleep monitoring,'' \emph{Sensors}, vol.~19, no.~9,
  p. 2133, 2019.

\bibitem{rhudy2014time}
M.~Rhudy, ``Time alignment techniques for experimental sensor data,''
  \emph{International Journal of Computer Science \& Engineering Survey
  (IJCSES)}, vol.~5, no.~2, pp. 1--14, April 2014.

\bibitem{zhang2017dynamic}
Z.~Zhang, R.~Tavenard, A.~Bailly, X.~Tang, P.~Tang, and T.~Corpetti, ``Dynamic
  time warping under limited warping path length,'' \emph{Information
  Sciences}, vol. 393, pp. 91--107, February 2017.

\bibitem{salvador2007toward}
S.~Salvador and P.~Chan, ``Toward accurate dynamic time warping in linear time
  and space,'' \emph{Intelligent Data Analysis}, vol.~11, no.~5, pp. 561--580,
  October 2007.

\bibitem{trajkovic2009modelling}
I.~Trajkovic, C.~Reller, M.~Wolf, and H.-A. Loeliger, ``Modelling and filtering
  almost periodic signals by time-varying fourier series with application to
  near-infrared spectroscopy,'' in \emph{2009 17th European Signal Processing
  Conference}.\hskip 1em plus 0.5em minus 0.4em\relax IEEE, 2009, pp. 632--636.

\bibitem{zhou2009canonical}
F.~Zhou and F.~Torre, ``Canonical time warping for alignment of human
  behavior,'' \emph{Advances in neural information processing systems},
  vol.~22, pp. 2286--2294, 2009.

\bibitem{zhou2012generalized}
F.~Zhou and F.~De~la Torre, ``Generalized time warping for multi-modal
  alignment of human motion,'' in \emph{2012 IEEE Conference on Computer Vision
  and Pattern Recognition}, IEEE.\hskip 1em plus 0.5em minus 0.4em\relax
  Providence, RI: IEEE, 2012, pp. 1282--1289.

\bibitem{rabiner1993fundamentals}
B.~J. L.~Rabiner, \emph{Fundamentals of Speech Recognition}.\hskip 1em plus
  0.5em minus 0.4em\relax River Street, Hoboken, NJ: PTR Prentice Hall, 1993,
  vol. 103.

\bibitem{mueen_extracting_2016}
\BIBentryALTinterwordspacing
A.~Mueen and E.~Keogh, ``Extracting {Optimal} {Performance} from {Dynamic}
  {Time} {Warping},'' in \emph{Proceedings of the 22nd {ACM} {SIGKDD}
  {International} {Conference} on {Knowledge} {Discovery} and {Data} {Mining}},
  ser. {KDD} '16.\hskip 1em plus 0.5em minus 0.4em\relax New York, NY, USA:
  Association for Computing Machinery, Aug. 2016, pp. 2129--2130. [Online].
  Available: \url{https://doi.org/10.1145/2939672.2945383}
\BIBentrySTDinterwordspacing

\bibitem{delong2012fast}
A.~Delong, A.~Osokin, H.~N. Isack, and Y.~Boykov, ``Fast approximate energy
  minimization with label costs,'' \emph{International journal of computer
  vision}, vol.~96, no.~1, pp. 1--27, 2012.

\bibitem{williams2020discovering}
A.~H. e.~a. Williams, ``Discovering precise temporal patterns in large-scale
  neural recordings through robust and interpretable time warping,''
  \emph{Neuron}, vol. 105, no.~2, pp. 246--259, 2020.

\bibitem{hotelling1992relations}
H.~Hotelling, ``Relations {Between} {Two} {Sets} of {Variates},'' in
  \emph{Breakthroughs in {Statistics}: {Methodology} and {Distribution}},
  S.~Kotz and N.~L. Johnson, Eds.\hskip 1em plus 0.5em minus 0.4em\relax New
  York, NY: Springer New York, 1992, pp. 162--190.

\bibitem{zhou_canonical_nodate}
F.~Zhou and F.~Torre, ``\BIBforeignlanguage{en}{Canonical time warping for
  alignment of human behavior},'' in \emph{\BIBforeignlanguage{en}{Advances in
  Neural Information Processing Systems}}, vol.~22.\hskip 1em plus 0.5em minus
  0.4em\relax Curran Associates, Inc., 2009, pp. 2286--2294.

\bibitem{trigeorgis_deep_2015}
G.~Trigeorgis, M.~A. Nicolaou, B.~W. Schuller, and Z.~Stefanos, ``Deep
  canonical time warping for simultaneous alignment and representation learning
  of sequences,'' \emph{IEEE Transactions on Pattern Analysis and Machine
  Intelligence}, vol.~40, no.~5, pp. 1128--1138, 2018.

\bibitem{watson1992characterization}
G.~A. Watson, ``Characterization of the subdifferential of some matrix norms,''
  \emph{Linear algebra and its applications}, vol. 170, pp. 33--45, 1992.

\bibitem{isack_energy-based_2012}
\BIBentryALTinterwordspacing
H.~Isack and Y.~Boykov, ``\BIBforeignlanguage{en}{Energy-{Based} {Geometric}
  {Multi}-model {Fitting}},'' \emph{\BIBforeignlanguage{en}{International
  Journal of Computer Vision}}, vol.~97, no.~2, pp. 123--147, Apr. 2012.
  [Online]. Available: \url{http://link.springer.com/10.1007/s11263-011-0474-7}
\BIBentrySTDinterwordspacing

\bibitem{boykov2001fast}
Y.~Boykov, O.~Veksler, and R.~Zabih, ``Fast approximate energy minimization via
  graph cuts,'' \emph{IEEE Transactions on pattern analysis and machine
  intelligence}, vol.~23, no.~11, pp. 1222--1239, 2001.

\bibitem{ARTSIG2021dataset}
\BIBentryALTinterwordspacing
A.~Botros, N.~Schütz, M.~A. Single, and T.~Nef, ``Artificial signal data for
  signal alignment testing.'' University of Bern, 2021. [Online]. Available:
  \url{https://zenodo.org/record/4522133}
\BIBentrySTDinterwordspacing

\bibitem{ECGSYN}
\BIBentryALTinterwordspacing
P.~E. McSharry and G.~D. Clifford, ``Ecgsyn - a realistic ecg waveform
  generator,'' Department of Engineering Science, University of Oxford and
  Laboratory for Computational Physiology, MIT, Oxford, UK and Massachusetts,
  USA, 2003. [Online]. Available: \url{https://physionet.org/content/ecgsyn/}
\BIBentrySTDinterwordspacing

\bibitem{McSharry2003dynamical}
P.~E. McSharry, G.~D. Clifford, L.~Tarassenko, and L.~A. Smith, ``A dynamical
  model for generating synthetic electrocardiogram signals.'' \emph{IEEE
  Transactions on Biomedical Engineering}, vol.~50, pp. 289--294, March 2003.

\bibitem{PhysioNet}
\BIBentryALTinterwordspacing
A.~e.~a. Goldberger, ``Physiobank, physiotoolkit , and physionet: Components of
  a new research resource for complex physiologic signals.''
  \emph{Circulation}, vol. 101, pp. 215--220, 2000. [Online]. Available:
  \url{https://physionet.org/}
\BIBentrySTDinterwordspacing

\bibitem{schmidt2018introducing}
P.~Schmidt, A.~Reiss, R.~Duerichen, C.~Marberger, and K.~Van~Laerhoven,
  ``Introducing wesad, a multimodal dataset for wearable stress and affect
  detection,'' in \emph{Proceedings of the 20th ACM international conference on
  multimodal interaction}, 2018, pp. 400--408.

\bibitem{botros2019long}
A.~e.~a. Botros, ``Long-term home-monitoring sensor technology in patients with
  parkinson’s disease—acceptance and adherence,'' \emph{Sensors}, vol.~19,
  no.~23, p. 5169, 2019.

\bibitem{lecun_misra}
\BIBentryALTinterwordspacing
Y.~LeCun and I.~Misra, ``Self-supervised learning: The dark matter of
  intelligence.'' [Online]. Available:
  \url{https://ai.facebook.com/blog/self-supervised-learning-the-dark-matter-of-intelligence/}
\BIBentrySTDinterwordspacing

\bibitem{2020SciPy-NMeth}
P.~e.~a. Virtanen, ``{{SciPy} 1.0: Fundamental Algorithms for Scientific
  Computing in Python},'' \emph{Nature Methods}, vol.~17, pp. 261--272, 2020.

\bibitem{harris2020array}
\BIBentryALTinterwordspacing
C.~R.~H. et~al., ``Array programming with {NumPy},'' \emph{Nature}, vol. 585,
  no. 7825, pp. 357--362, Sep. 2020. [Online]. Available:
  \url{https://doi.org/10.1038/s41586-020-2649-2}
\BIBentrySTDinterwordspacing

\bibitem{mckinney2010data}
W.~McKinney \emph{et~al.}, ``Data structures for statistical computing in
  python,'' in \emph{Proceedings of the 9th Python in Science Conference}, vol.
  445.\hskip 1em plus 0.5em minus 0.4em\relax Austin, TX, 2010, pp. 51--56.

\bibitem{NEURIPS2019_9015}
\BIBentryALTinterwordspacing
A.~e.~a. Paszke, ``Pytorch: An imperative style, high-performance deep learning
  library,'' in \emph{Advances in Neural Information Processing Systems 32},
  H.~Wallach, H.~Larochelle, A.~Beygelzimer, F.~d\textquotesingle
  Alch\'{e}-Buc, E.~Fox, and R.~Garnett, Eds.\hskip 1em plus 0.5em minus
  0.4em\relax Curran Associates, Inc., 2019, pp. 8024--8035. [Online].
  Available:
  \url{http://papers.neurips.cc/paper/9015-pytorch-an-imperative-style-high-performance-deep-learning-library.pdf}
\BIBentrySTDinterwordspacing

\bibitem{boykov2004experimental}
Y.~Boykov and V.~Kolmogorov, ``An experimental comparison of min-cut/max-flow
  algorithms for energy minimization in vision,'' \emph{IEEE transactions on
  pattern analysis and machine intelligence}, vol.~26, no.~9, pp. 1124--1137,
  2004.

\bibitem{escorihuela2020reduced}
R.~M. Escorihuela, L.~Capdevila, J.~R. Castro, M.~C. Zaragoz{\`a}, S.~Maurel,
  J.~Alegre, and J.~Castro-Marrero, ``Reduced heart rate variability predicts
  fatigue severity in individuals with chronic fatigue syndrome/myalgic
  encephalomyelitis,'' \emph{Journal of translational medicine}, vol.~18,
  no.~1, pp. 1--12, 2020.

\end{thebibliography}
\newpage

\section*{Appendix}
\setcounter{figure}{0}
\setcounter{section}{1}
\setcounter{table}{0}
\renewcommand\thefigure{\Alph{section}.\arabic{figure}}
\renewcommand\thetable{\Alph{section}.\Roman{table}}

\subsection*{Code and Data Availability}
For non-commercial applications, our implementation may be obtained upon request.
Please contact one of the authors directly.
The remaining data used is either public, where the sources are referenced, or private and protected by applicable Swiss regulations.
Since sharing private medical data would require a respective waiver from the ethics committee, we can only provide those upon personal request and under suitable conditions, please contact the last author in this case.

\subsection{Hyperparameters}
\subsubsection*{Segmentation}
Windows sizes $w$ for all datasets are given in table~\ref{table:hyperparameters}, together with the sampling frequency of the datasets.
Choosing window length $w$ thus comes with three main considerations, 1) $w$ has to be large enough to contain sensor offsets and clock-drifts within a given super-segment, and 2) $w$ has to be small enough, such that drift in a single segment is negligible.
For all $w$ that satisfy consideration 1) and 2), the optimal value becomes a trade-off between alignment accuracy and computational speed, where shorter windows reduce the constant computational overhead while longer ones can improve accuracy, especially with almost periodic signals that only exhibit minimal inter-cycle variability.
In practice, for most sensors with $f_s \gg 1Hz$, a reasonable value can be anywhere between 10 seconds and 3 minutes, but this depends on the super-segment length and each pair of sensors. 
In terms of computational complexity $w$ influences the number of segments $M$ (assuming that super-segment length $z$ is held constant).
As such, one might consider the computational complexity of the reconstruction stage, which is directly affected by $M$.
According to \cite{delong2012fast}, the energy based model extraction with $\alpha$-expansion  has a complexity of $\mathcal{O}(LN)$, where $L$ denotes the number of models we want to fit. 
For a single super-segment the complexity is therefore $\mathcal{O}(LM)$. 
Thus, one does generally not want super-segments that contain too many segments.
In practice, for most real sensors, we found super-segment lengths of up to two weeks to be reasonable, but obviously this depends on how quickly a given pair of sensors drift from each other.
An additional practical consideration, if one runs DCCA on a GPU, is the GPU memory which may become a bottleneck for very long $w$, especially with multi-variate signals.

\subsubsection*{Neural Networks}
\begin{figure}
\centering
\subfloat[]{\includegraphics[width=0.75\columnwidth]{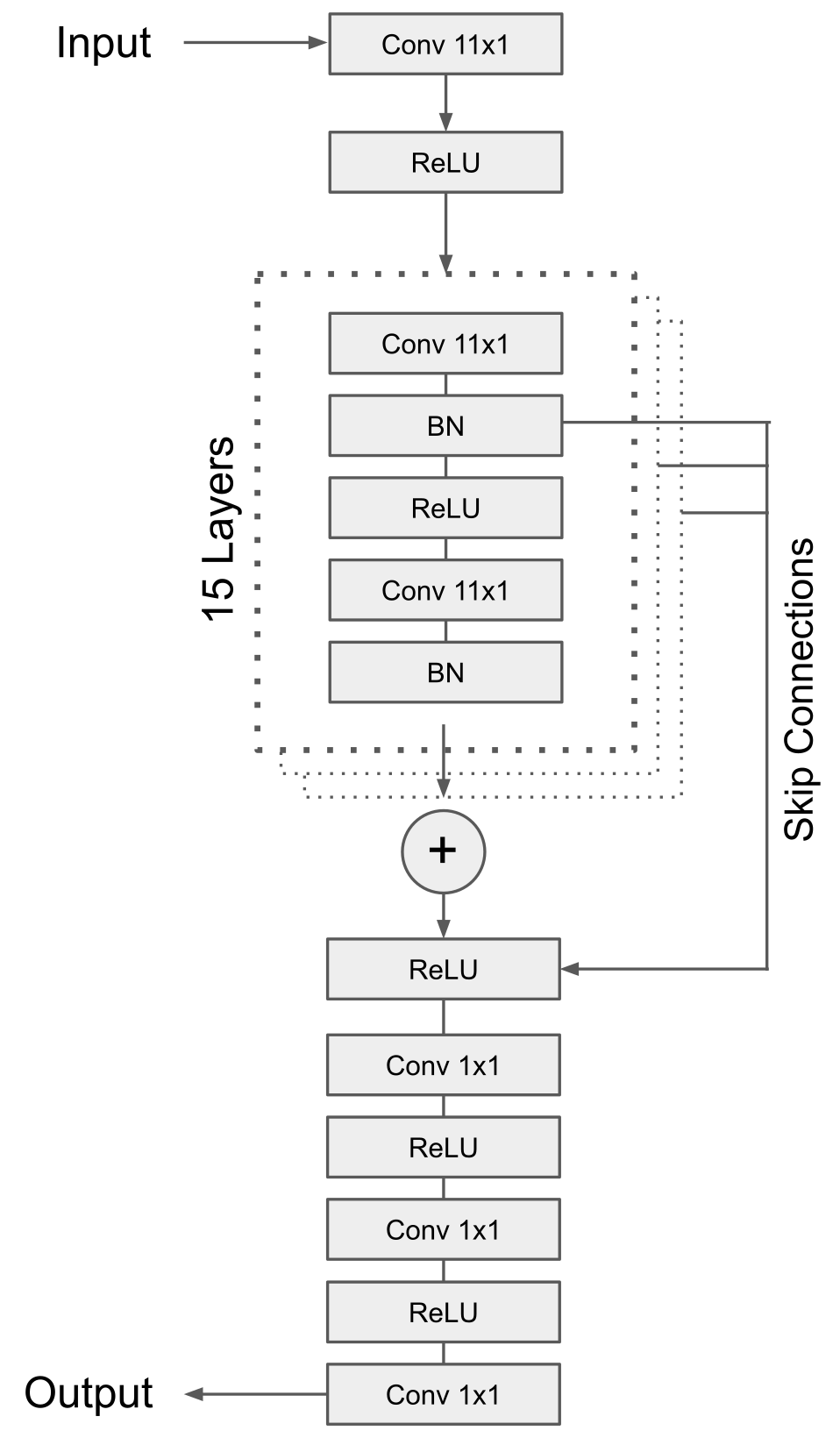}%
}
\caption{Depicted is the model used to approximate the transformation functions in DCCA - a regular 1D Convolutional Residual Neural Network. For all experiments and shown examples the same parameterization (except for input/output channels) was used. The kernel size was set to 11 for all convolutional layers, stride was set to 1 and padding to same.}
\label{fig:resnet}
\end{figure}

We found a one-dimensional residual Convolutional Neural Network (CNN) architecture, with locally constrained receptive fields, to work well for the alignment task.
It allowed for effective computations and adjustments of network capacity based on the network depth. 
While the network architecture is not crucial here, we found it highly important that the receptive field of the CNN is not too large, as it otherwise tends to internally shift signals, which would lead to incorrect alignments. 
The network architecture is described in Fig.~\ref{fig:resnet}.
The same hyperparameters were used across all shown experiments and examples.
Kernel size was set to 11, with stride 1 and same padding. 
As nonlinearity we used leaky rectified linear units. 
Optimization was performed with the commonly used Adam optimizer and an initial learning rate of 0.0004.
The number of full passes over the full dataset, thus the number of epochs, was set 25.
Area penalty governing hyperparameter $\beta$ was set to 0.1.
Pre-training was only applied in the self-supervised proof of concept study, to keep variation between multiple alignments as small as possible.

\subsubsection*{Correction Model Extraction} 
A detailed explanation of the effects of these hyper parameters on the model calculations is given in~\cite{isack_energy-based_2012}.
For the PE{\footnotesize A}RL-algorithm, there are three relevant hyper parameters we have set.
The model penalty parameter $h_L$ is a regularization term for the number of different models chosen by the algorithm. 
We set this parameter to $h_L = 50$.
The parameter $c$ is used to weight the shape of the neighbourhood of points. 
It has a direct influence on the smoothness of the models.
We set this parameter to $c = 10$.
$\gamma$ provides a threshold for inlier models.
This parameter is a cost value assigned to all points deemed to be outliers. 
It affects the area of effect for all discovered models.
We set this parameter to $\gamma = 10$.
The family of extraction functions was set to be quadratic.
Throughout all experiments we used this parametrization, which proved sufficient.

\begin{table*}
\caption{Hyperparameters window size $w$ and sampling frequency $f_s$ for all used datasets.}
\label{table:hyperparameters}
\scriptsize
\begin{tabular}{lrrcc}
    \toprule
    & $w$ & $f_s$ & \textbf{Sensor1/ Placement} & \textbf{Sensor2/ Placement} \\
    \midrule
Baseline & 50005ms             & 200 Hz                 & Artifical/ -              & Artifical/ -                \\ \hline
Poisson 1D & 50010ms              & 100 Hz               & Artifical/ -                & Artificial/ -                 \\ \hline
Poisson 3D & 50005ms              & 200 Hz                 & Artifical/ -                 & Artificial/ -                  \\ \hline
ECG-BCG & 50005ms              & 200 Hz                 & Artifical/ -                 & Artificial/ -                  \\ \hline
ECG-ECG 1 & 15001ms              & 1000 Hz                 & NeXus-10 MKII ECG/ Chest             & eMotion Faros 360° ECG/ Chest               \\ \hline
ECG-ECG 2 & 15004ms              & 500 Hz                 & NeXus-10 MKII ECG/ Chest              & SOMNOtouch NIBP ECG/ Chest               \\ \hline
PPG-ECG & 15004ms              & 250 Hz                  & NeXus-10 MKII ECG/ Chest              & SOMNOtouch NIBP PPG/ Chest               \\ \hline
aVL-ECG & 15004ms              & 500 Hz                 & NeXus-10 MKII ECG/ Chest              & SOMNOtouch NIBP ECG aVL-Lead/ Chest              \\ \hline
ACC-ECG & 60004ms              & 250 Hz                 & NeXus-10 MKII ECG/ Chest              & eMotion Faros 360° 3-axis Accelerometer/ Upper Body \\ \hline
Fig. 5 a) & 60020ms              & 50 Hz          & EMFIT QS Low-Frequency Band/ Beneath Mattress  &  Carescape Monitor B650 Respiration/ Chest \\ \hline
Fig. 5 b) & 30020ms              & 50 Hz          & RespiBAN ECG/ Chest                     & Empatica E4 PPG/ Wrist \\ \hline
Fig. 5 c) & 30020ms              & 50 Hz          & Jaeger-Toennies EEG (C3-A2)/ Head        & Jaeger-Toennies ECG/ Chest \\ \hline
Fig. 5 d) & 120010ms              & 100 Hz          & Axivity AX3 3-Axis Accelerometer/ wrist        & Axivity AX3 3-Axis Accelerometer/ Ankle \\ \hline
Sec. VI & 30010ms              & 100 Hz          & EMFIT QS High-Frequency Band/ Beneath Mattress        & Carescape Monitor B650 ECG \\
\bottomrule
\end{tabular}
\end{table*}

\subsection*{Details on Experiment Datasets}
Regarding the simultaneous sensor recordings dataset by Vollmer et al. \cite{vollmer2019alignment}, we only used records from the participants 5-13, as the initial recordings had slightly different formats in some cases.
Additional details with respect to each dataset are shown in Table \ref{table:datasets}.

\begin{table*}
\centering
\scriptsize
\caption{Dataset details, showing the exact records we used for experiments as well the the respective signal indices (if not the full signal was used), as well as average signal length of each record of a dataset.}
\label{table:datasets}
\begin{tabular}{lllll}
\toprule
Dataset & Record Identifier & Indices & Average Length \\
\midrule
Baseline   & N/A            & Full Signal       & 10 h      \\ \hline
Poisson 1D & 1:6            & Full Signal       & 12 h      \\ \hline
Poisson 3D & 1:5            & Full Signal       & 12 h      \\ \hline
ECG-BCG    & 0:5            & 0:999'999          & 10.8 h    \\ \hline
ECG-ECG 1  & 5:13           & Full Signal       & 30 min    \\ \hline
ECG-ECG 2  & 5:13           & Full Signal       & 30 min    \\ \hline
PPG-ECG    & 5:13           & Full Signal       & 30 min    \\ \hline
aVL-ECG    & 5:13           & Full Signal       & 30 min    \\ \hline
ACC-ECG    & 5:13           & Full Signal       & 30 min \\
\bottomrule
\end{tabular}
\end{table*}

\subsection*{Implementation}
The algorithm shown was implemented using Python 3.7, \textit{SciPy} v1.4.1 \cite{2020SciPy-NMeth}, \textit{NumPy} v1.17.0 \cite{harris2020array}, \textit{Pandas} v1.1.5 \cite{mckinney2010data}, \textit{PyTorch} v1.7.0 \cite{NEURIPS2019_9015} and for the energy based multi-model extraction algorithm we used the \textit{gco3.0} library \cite{boykov2004experimental, boykov2001fast, delong2012fast}.

\subsection*{Synthetic Data}
All synthetic datasets are available online at https://doi.org/10.5281/zenodo.4522133.
The ECGSYM software used for the generation of ECG data is available at~\cite{ECGSYN, PhysioNet}.

\subsection*{Details on Self-Supervised Learning Proof of Concept Study}\label{app:sec_self_supervised}
\begin{figure}
\subfloat[]{\includegraphics[width=\columnwidth]{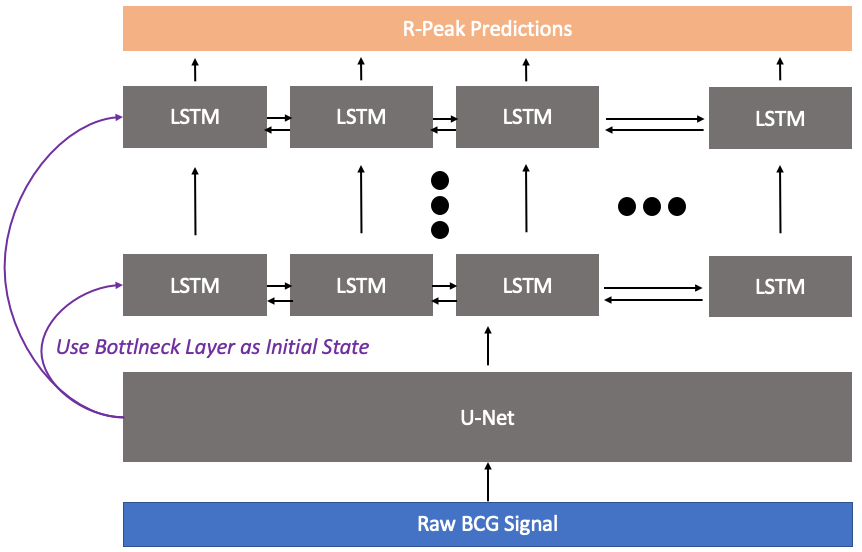}%
}
\caption{Depicts the U-Net Bidirectional Long Short Term Memory Model, used in the Self-Supervised Proof of Concept Study. The U-Net consists of 5 blocks with a kernel size of 7, while the bidirectional LSTM has three hidden layers, dropout of 0.3 and is initialized with the bottleneck layer of the U-Net model as initial state.}
\label{fig:lstm}
\end{figure}

The devices used for this example were an EMFIT QS (Emfit Ltd) for the BCG signal and a Carescape Monitor B650 (GE Healthcare, Little Chalfont, UK) for the 5-lead ECG recording.
The data was collected in the context of a study examining how uni- versus multi-modal stimuli affects the relaxing effect of virtual reality. 
Participants were healthy adults (25 female, 17 male), aged between 24 and 83 years old (mean 60.2, SD 15.3).
We made sure to randomly split training and test sets based on participants, so as not to leak any information.
For R-peak extraction on the basis of the model output we used the SciPy \cite{2020SciPy-NMeth} function "\textit{find\_peaks}", where peaks must have a minimal height of 0.05 (after squashing the output of the last layer through a sigmoid) and be spaced at least 330ms apart.
Additionally, segments resulting in unrealistic heart-rates below 40 or above 160 were filtered out.
The employed neural network model is depicted and described in Fig.~\ref{fig:lstm}.
The idea was to use U-Net for R-peak segmentation and let the Bidirectional LSTM handle temporal dynamics, such as realistic spacing between regular heart beats.
To provide the LSTM with a global context of a segment we additionally initialzed its initial state with the bottleneck-layer of the U-Net.
We did some preliminary testing of this architecture against other alternatives (such as only U-Net, only LSTM, WaveNet inspired models as well as vanilla CNN models), but no thorough evaluation was done, as we mainly wanted to show that this kind of scenario can work after DCCA alignment.

\newpage

\subsection*{Alignment of almost periodic signals}\label{app:sec_almost_periodic}
\begin{figure*}[!t]
\subfloat[]{\includegraphics[width=0.5\columnwidth]{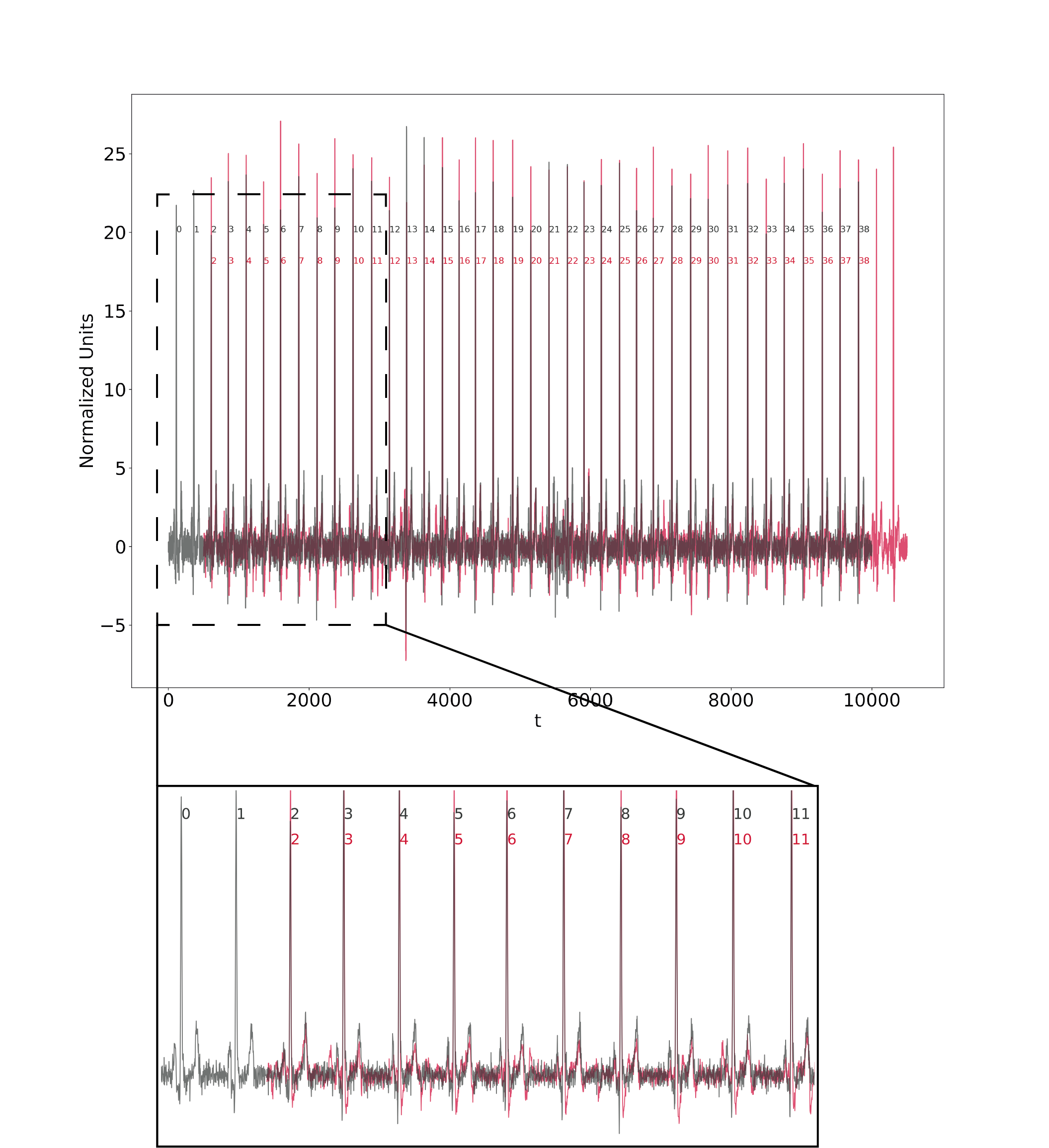}%
\label{fig:periodicalignmentcorrect}}
\hfil
\subfloat[]{\includegraphics[width=0.5\columnwidth]{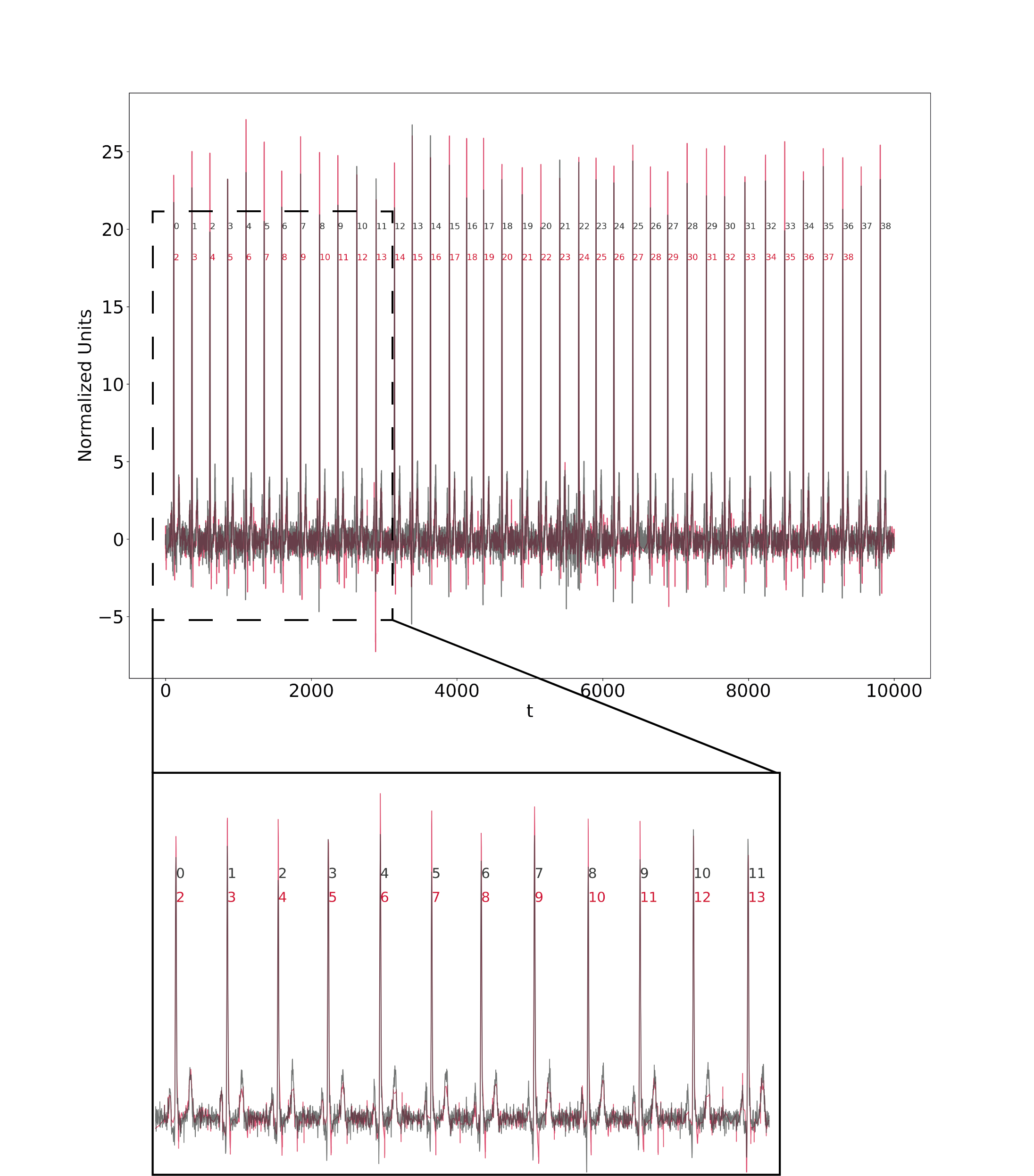}%
\label{fig:periodicalignmentwrong}}
\hfil
\subfloat[]{\includegraphics[width=0.5\columnwidth]{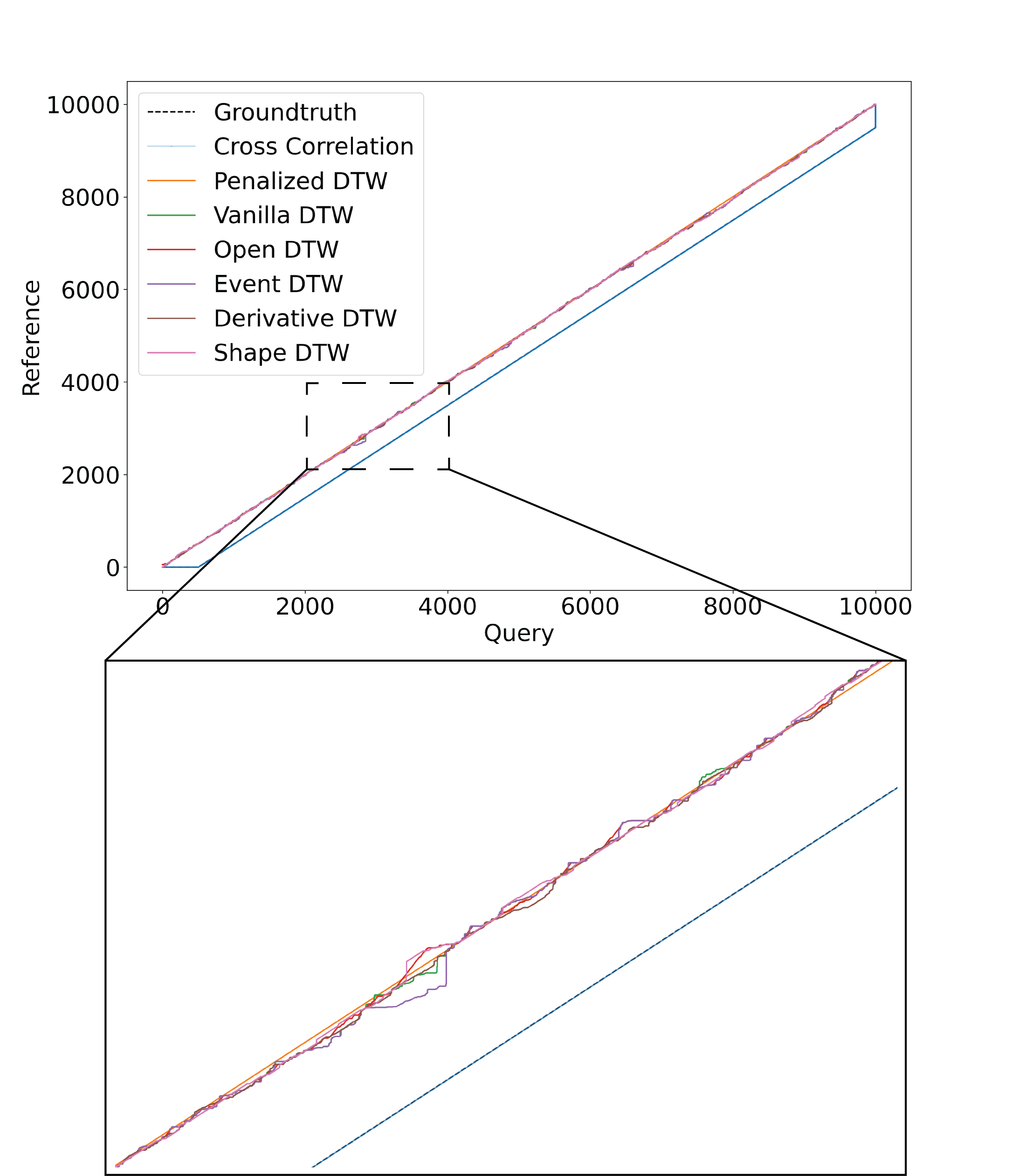}%
\label{fig:warpingpath}}
\caption{
Alignment of two ECG signals, showcasing the problem of non-linear warping with almost periodic signals. 
The numbers indicate the correct heart-beat enumerations.
The red signal is missing the first two heart-beats.
In (a), a normalized cross-correlation is used for alignment, which results in a shift by two heart-beats, which is the correct alignment.
In (b), DTW-based algorithms are used for the alignment, which automatically corrects for small variations in the periodicity.
As a result, the two signals are aligned with the closest peak. 
This, however, leads to an incorrect alignment as can be seen by the enumeration of the respective peaks.
In (c), the alignment paths are depicted in matrix form.
Only the cross-correlation finds the correct offset of two heart-beats while the DTW-based algorithms only find alignments close to the diagonal, as it eliminates small inter-cycle variations.
}
\label{fig:alignmen_tDTW_CC}
\end{figure*}

Almost periodic signals are signals that are overall periodic, but with a certain margin or error
\begin{equation}
    |S(t+T) - S(t)| < \epsilon
\end{equation}
with $\epsilon > 0$.
Trajkiciv et al.  state that many biomedical signals are almost periodic~\cite{trajkovic2009modelling}.
An example is the heart rate, that is - at rest - almost constant.
But an important factor for medical assessment is the heart-rate-variability (HRV), an expression for how much the heart-rate shifts around its major frequency and is also denoted \textit{cycle length variability}.
In general, lower HRV is linked to a variety of adverse health effects~\cite{escorihuela2020reduced}.
When aligning two signals monitoring the heart beat, this variability can help the correct alignment.
This is shown in Fig.~\ref{fig:alignmen_tDTW_CC}, where two ECG signals are aligned, using CC in Fig.~\ref{fig:periodicalignmentcorrect} and multiple DTW approaches in Fig.~\ref{fig:periodicalignmentwrong}.
Note that the same is true for a majority of biological (or naturally occurring) signals.

\end{document}